\def\year{2022}\relax
\documentclass[letterpaper]{article} 
\usepackage{aaai22}  
\usepackage{times}  
\usepackage{helvet}  
\usepackage{courier}  
\usepackage[hyphens]{url}  
\usepackage{graphicx} 
\usepackage{natbib}  
\usepackage{caption} 
\usepackage{acro}
\usepackage{epsfig}
\usepackage{amsmath}
\usepackage{amssymb}
\usepackage{bm}
\usepackage{upgreek}
\usepackage[utf8x]{inputenc} 
\usepackage{array}

\DeclareCaptionStyle{ruled}{labelfont=normalfont,labelsep=colon,strut=off} 
\usepackage{algorithm}
\usepackage{algorithmic}

\usepackage{newfloat}
\usepackage{listings}
\floatstyle{ruled}
\newfloat{listing}{tb}{lst}{}
\floatname{listing}{Listing}
\title{Deep Implicit Statistical Shape Models for 3D Medical Image Delineation}
 \author {
     Ashwin Raju \textsuperscript{\rm{1,2}}
     Shun Miao \textsuperscript{\rm 1}
     Dakai Jin \textsuperscript{\rm 1}
     Le Lu \textsuperscript{\rm 1}
     Junzhou Huang \textsuperscript{\rm 2}
     Adam P. Harrison \textsuperscript{\rm 1}\thanks{Corresponding Author}
 }
 \affiliations {
     \textsuperscript{\rm 1} PAII Inc, Bethesda, MD, USA\\
     \textsuperscript{\rm 2} University of Texas at Arlington, Arlington, TX, USA\\
     \{ashwin.raju93, jzhuang\}@uta.edu \\
     \{shwinmiao, dakai.jin, tiger.lelu, adam.p.harrison\}@gmail.com
     
 }

\DeclareAcronym{PnP}{
short=PnP,
long=Perspective-n-Point
}
\DeclareAcronym{MLP}{
short=MLP,
long=multilayer perceptron
}

\DeclareAcronym{DoF}{
short=DoF,
long=degrees of freedom
}
\DeclareAcronym{SDF}{
short=SDF,
long=signed distance function
}

\DeclareAcronym{ASM}{
short=ASM,
long=active shape model
}
\DeclareAcronym{CT}{
short=CT,
long=computed tomography
}

\DeclareAcronym{HCC}{
short=HCC,
long=hepatocellular carcinoma
}

\DeclareAcronym{ICC}{
short=ICC,
long=intrahepatic cholangiocellular carcinoma
}

\DeclareAcronym{TACE}{
short=TACE,
long=transarterial chemoembolization
}

\DeclareAcronym{SSM}{
short=SSM,
long=statistical shape model
}
\DeclareAcronym{MSL}{
short=MSL,
long=marginal space learning
}
\DeclareAcronym{RL}{
short=RL,
long=Deep Reinforcement learning
}

\DeclareAcronym{DSC}{
short=DSC,
long=Dice-S\o{}rensen coefficient,
foreign-plural={}
}
\DeclareAcronym{HD}{
short=HD,
long=Hausdorff distance,
foreign-plural={}
}

\DeclareAcronym{OAR}{
short=OAR,
long=organ-at-risk
}

\DeclareAcronym{MSD}{
short=MSD,
long=medical segmentation decathlon 
}

\DeclareAcronym{PCA}{
short=PCA,
long=principle components analysis 
}

\DeclareAcronym{DRL}{
short=DRL,
long=deep reinforcement learning 
}

\DeclareAcronym{DIAS}{
short=DISSM,
long=deep implicit statistical shape model
}

\DeclareAcronym{PDM}{
short=PDM,
long=point distribution model 
}

\DeclareAcronym{FCN}{
short=FCN,
long=fully-convolutional network 
}

\DeclareAcronym{CNN}{
short=CNN,
long=convolutional neural network 
}

\DeclareAcronym{MDP}{
short=MDP,
long=Markov decision process 
}

\DeclareAcronym{AE}{
short=AE,
long=auto-encoder
}
\DeclareAcronym{ASSD}{
short=ASSD,
long=average symmetric surface distance
}
\DeclareAcronym{GAN}{
short=GAN,
long=generative adversarial network
}
\DeclareAcronym{HU}{
short=HU,
long=Hounsfield units
}

\newcommand{\Tau}{\mathrm{T}}
\newcommand{\SDF}{\mathit{SDF}}
\newcommand{\SDFmu}{\mathit{SDF}_{\boldsymbol\upmu}}
\newcommand{\SDFpca}{\mathit{SDF}_{\boldsymbol\lambda}}
\newcommand{\renderwidth}{.15\linewidth}
\newcommand{\stageswidth}{.25\linewidth}
\newcommand{\qualwidth}{2.17cm}
\newcommand{\qualheight}{1.76cm}

\DeclareMathOperator*{\argmin}{arg\,min}
\newcommand{\eg}{\textit{e}.\textit{g}.}
\newcommand{\ie}{\textit{i}.\textit{e}.}
\newcommand{\etal}{\textit{et al}. }

\begin{document}

\maketitle
\renewcommand{\thepage}{S\arabic{page}} 
\renewcommand{\thesection}{S\arabic{section}}  
\renewcommand{\thetable}{S\arabic{table}}  
\renewcommand{\thefigure}{S\arabic{figure}}
\begin{abstract}
3D delineation of anatomical structures is a cardinal goal in medical imaging analysis. Prior to deep learning, \acp{SSM} that imposed anatomical constraints and produced high quality surfaces were a core technology. Today's \acp{FCN}, while dominant, do not offer these capabilities. We present \acp{DIAS}, a new approach that marries the representation power of deep networks with the benefits of \acp{SSM}. \acp{DIAS} use an implicit representation to produce compact and descriptive deep surface embeddings that permit statistical models of anatomical variance. To reliably fit anatomically plausible shapes to an image, we introduce a novel rigid and non-rigid pose estimation pipeline that is modelled as a \ac{MDP}.  Intra-dataset experiments on the task of pathological liver segmentation demonstrate that \acp{DIAS} can perform more robustly than four leading \ac{FCN} models, including nnU-Net + an adversarial prior: reducing the mean \ac{HD} by $7.5$-$14.3$ mm and improving the \emph{worst case} \ac{DSC} by $1.2$-$2.3\%$. More critically, cross-dataset experiments on an external and highly challenging clinical dataset demonstrate that \acp{DIAS} improve the \emph{mean} \ac{DSC} and \ac{HD} by $2.1$-$5.9\%$ and $9.9$-$24.5$ mm, respectively, and the \emph{worst-case} \ac{DSC} by $5.4$-$7.3\%$. Supplemental validation on a highly challenging and low-contrast larynx dataset further demonstrate \ac{DIAS}'s improvements. These improvements are over and above any benefits from representing delineations with high-quality surfaces. 
\end{abstract}

\acresetall

\section{Introduction}

\begin{figure}
    \centering
    
         \includegraphics[width=.7\linewidth]{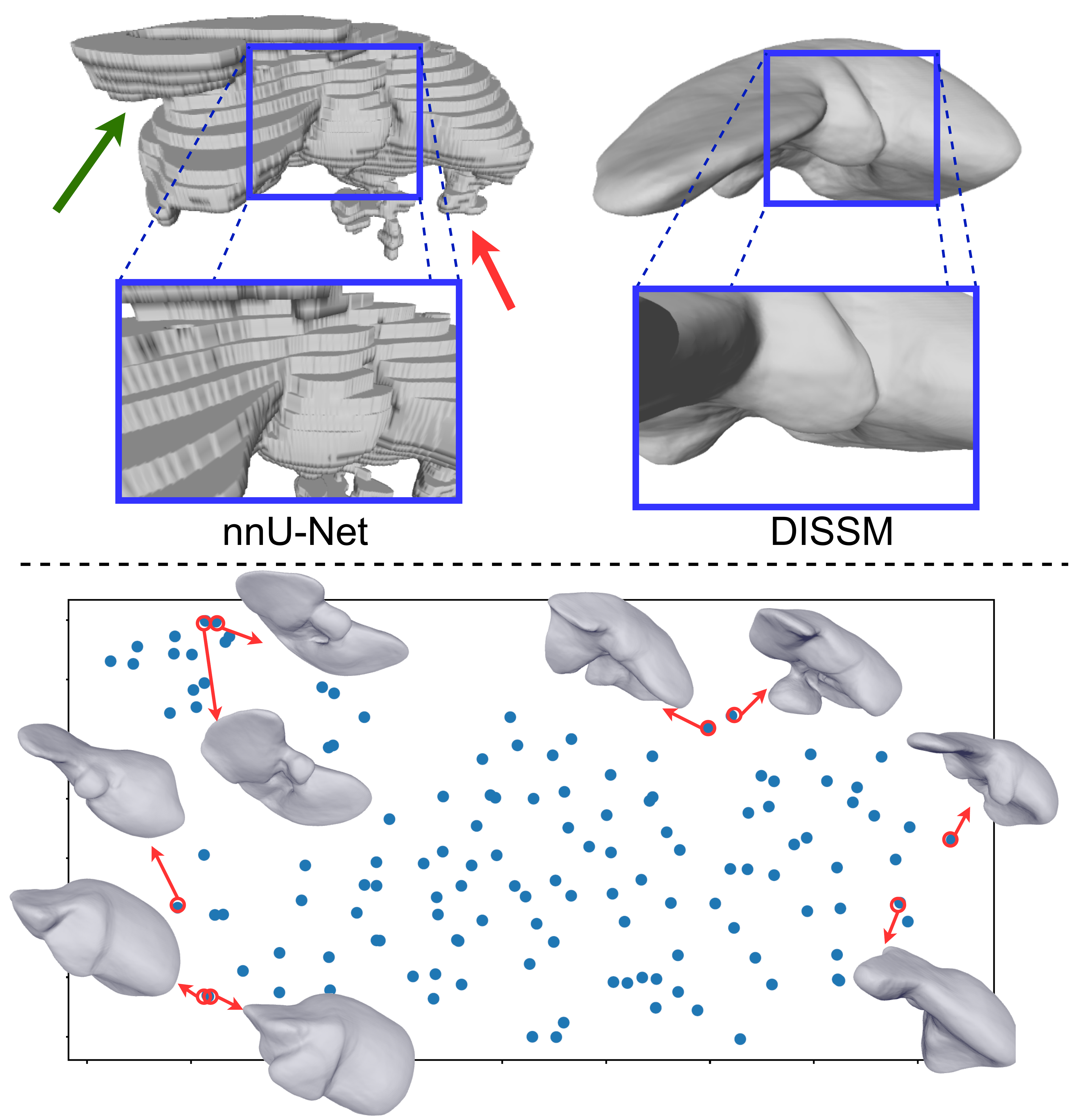}
    
    \caption{\textbf{Top Panel}: \acsp{DIAS} produce high quality surfaces without discretization. Its use of rich and explicit anatomical priors ensure robustness, even on highly challenging cross-dataset clinical samples. In this example, nnU-Net~\cite{isensee_nnu-net_2021} oversegments the cardiac region (green arrow) and mishandles a \acs{TACE}-treated lesion (red arrow), causing a fragmented effect. \textbf{Bottom Panel}: a 2D t-SNE embedding~\cite{Maaten_2008} of the \acs{DIAS} shape latent space. Shapes closer together share similar features. }
    \label{fig:intro}
\end{figure}

3D delineation is a fundamental task in medical imaging analysis. Currently, medical segmentation is dominated by \acp{FCN}~\cite{Long_2015_CVPR},  which segment each pixel or voxel in a bottom-up fashion. \acp{FCN} are well-suited to the underlying \ac{CNN} technology and are  straightforward to implement using modern deep learning software. The current state has been made particularly plain by the dominance of  nnU-Net~\cite{isensee_nnu-net_2021}, \ie{}, nothing-new U-Net, in the \ac{MSD} challenge~\cite{Simpson_2019}. Yet, despite their undisputed abilities, \acp{FCN} do lack important features compared to prior technology. For one, a surface-based representation is usually the desired end product, but \acp{FCN} output masks, which suffer from discretization effects (see top panel of Fig.~\ref{fig:intro}). These are particularly severe when large inter-slice distances come into play. Conversion to a smoothed mesh is possible, but it introduces its own artifacts. While this is an important drawback, an arguably more critical limitation is that current \ac{FCN} pipelines typically operate  with no shape constraints.

Shape priors are critical for ensuring anatomically plausible delineations, but the techniques and concepts of \acp{SSM}, so integral prior to deep learning~\cite{Heimann_2009}, have fallen out of favor.  Despite the incontrovertible power of \acp{FCN}, they can  produce egregious mistakes, especially when presented with morbidities, scanners, or other scenarios not seen in training (again see top panel of Fig.~\ref{fig:intro}). Because it is impossible to represent all clinical scenarios well enough in training datasets, priors can act as valuable regularizing forces. \acp{FCN} may also struggle with anatomical structures with low-contrast boundaries. Efforts have been made to incorporate anatomical priors with \acp{CNN}, but these either do not directly model shape and/or do not estimate rigid poses. Thus, they do not construct a true \ac{SSM}. Even should their interoperation with \acp{CNN} not be an issue, classic \acp{SSM} also have their own disadvantages, as they mostly rely on the \ac{PDM}~\cite{Cootes_1992}, which requires determining a correspondence across shapes. Ideally, correspondences would not be required. 


To fill these gaps, we introduce \acp{DIAS}, a new and deep approach to \acp{SSM}. Using the recently introduced deep implicit shape concept~\cite{park2019deepsdf, chen_learning_2019,mescheder_occupancy_2019}, \acp{DIAS} learn a compact and rich latent space that can accurately and densely generate the \acp{SDF} of a representative sample of shapes. Importantly, correspondence between shapes is  unnecessary, eliminating a major challenge with traditional \acp{SSM}. Statistics, \eg, mean shape, \ac{PCA}, and interpolation, can be performed directly on the latent space (see bottom panel of Fig.~\ref{fig:intro}). To fit an anatomically plausible shape to a given image, \acp{DIAS} use a \ac{CNN} to determine rigid and non-rigid poses, thus marrying the representation power of deep networks with anatomically-constrained surface representations. Pose estimation is modelled as a \ac{MDP} to determine a trajectory of rigid and non-rigid poses, where the latter are defined as \ac{PCA} loadings of the \ac{DIAS} latent space. To handle the intractably large search space of poses, \acp{DIAS} make use of \ac{MSL}~\cite{Zheng_2008,zheng2014marginal} and inverted episodic training, the latter a concept we introduce. A final constrained deep level set refinement~\cite{michalkiewicz_implicit_2019} captures any fine details not represented by \ac{DIAS} latent shape space. At a high level, \acp{DIAS} share many philosophies with traditional \acp{SSM}, but they modernize these concepts in a powerful deep-learning framework.

As proof of concept, we evaluate \ac{DIAS} primarily on the problem of 3D pathological liver delineation from \ac{CT} scans, with supplemental validation on a challenging larynx segmentation task from low-contrast \ac{CT}. For the former, we compare our approach to leading 2D~\cite{Harrison_2017}, hybrid~\cite{Li_2018}, 3D cascaded~\cite{isensee_nnu-net_2021}, and adversarial learning~\cite{yang_automatic_2017} \acp{FCN}. When trained and tested on the \ac{MSD} liver dataset~\cite{Simpson_2019}, \acp{DIAS} provide more robust delineations, improving the mean \ac{HD} by $7.5$-$14.3$mm. \emph{This is over and above any benefits of directly outputting a high resolution surface.} More convincingly, we perform cross-dataset evaluation on an external dataset ($97$ volumes) that directly reflect  clinical conditions~\cite{Raju_2020}. \acp{DIAS} improve the mean \ac{DSC} and \ac{HD} from $92.4\%$ to $95.9\%$ and from $34.1$mm to $21.8$mm, respectively, over the best fully \ac{FCN} alternative (nnU-Net). In terms of robustness, the worst-case \ac{DSC} is boosted from $88.1\%$ to $93.4\%$. Commensurate improvements are also observed on the larynx dataset. These results confirm the value, once taken for granted, of incorporating anatomical priors in 3D delineation. Our contributions are thus: 1) we are the first to introduce a true \emph{deep} \ac{SSM} model that outputs high resolution surfaces; 2) we build a new correspondence-free, compact, and descriptive anatomical prior; 3) we present a novel pose estimation scheme that incorporates inverted episodic training and \ac{MSL}; and 4) we provide a more robust solution than leading \acp{FCN} in both intra- and cross-dataset evaluations. Our code is publicly shared\footnote{\url{https://github.com/AshStuff/dissm}}.

\section{Related Work} \label{related_works}


\noindent {\bf Anatomical Shape Priors}: Because anatomical structures are highly constrained, delineations should match shape priors. \acp{SSM} were a workhorse for 3D medical segmentation and were most popularly realized as \acp{PDM}~\cite{Cootes_1992}, which requires determining a set of surface point correspondences across all shapes. Correspondence permitted statistical techniques, such as \ac{PCA}, to model shape variability~\cite{Heimann_2009}.  However, dense 3D surface landmarks are not typically possible to define, and even when so, they are not always reliably present~\cite{Heimann_2009}. The alternative of automatically generating correspondences is still a dense research topic~\cite{Heimann_2009}, and all existing solutions are imperfect. While \acp{PDM} have been used with \acp{CNN}~\cite{milletari_integrating_2017,bhalodia_deepssm_2018}, these works make no attempt at determining rigid poses, thus they are only applicable in constrained and limited setups. In general the reliance on explicit surface representations, \eg{} meshes, makes it difficult to integrate traditional \acp{SSM} with \acp{CNN}. \acp{SSM} based on implicit level set representations have been explored~\cite{cremers_review_2007}, but the statistics must be collected across a dense 3D regular grid. The most popular approach today is to impose an implicit \emph{prior} (different from implicit \emph{representation}) using auto-encoders~\cite{oktay_anatomically_2017,ravishankar_learning_2017} or generative adversarial networks~\cite{yang_automatic_2017,raju_user-guided_2020,cai_end--end_2019}, but these do not permit a controllable and interpretable prior and are not invariant to rigid similarity transforms, so a true shape prior is not constructed. Recent work has also used \acp{CNN} to deform an initial mesh~\cite{yao_integrating_2019,wickramasinghe_voxel2mesh_2020} or point-cloud~\cite{cai_end--end_2019} sphere, but the process is used to augment an existing voxel-based output, otherwise the deformation cannot capture all details~\cite{wickramasinghe_voxel2mesh_2020}. Besides, these works offer no statistical description of anatomical priors. Like traditional \acp{SSM}, \acp{DIAS} explicitly enforce shape constraints via a model of anatomical variability, but using a deep implicit \ac{SDF} \emph{representation} that is 1) highly descriptive, 2) highly compact, and 3) requires no correspondences. \ac{DIAS} is the first to offer a complete ``deep'' \ac{SSM} solution that includes rigid and non-rigid pose estimation.

\noindent {\bf Deep Implicit Shapes}: The concept of deep implicit shapes~\cite{park2019deepsdf, chen_learning_2019,mescheder_occupancy_2019} were introduced to represent shapes using a compact latent space and an \ac{SDF} decoder. \acp{DIAS} adopt \citet{park2019deepsdf}'s auto-decoder formulation. Unlike these seminal works, \acp{DIAS} build a model of anatomical variability and propose a pose estimation strategy for 3D images. 

\noindent {\bf Pose Estimation}: Fitting an \ac{SSM} to an image requires determining the rigid and non-rigid poses of the shape. Standard learning strategies, such as one shot regression~\cite{vstern2016local,gauriau2015multi} and exhaustive scanning~\cite{zheng20153d}, yield sub-optimal results as the model relies on a prohibitively large number of samples~\cite{ghesu2017multi} and only has one opportunity to produce pose estimations. A recent work directly regresses poses using deep implicit shapes, but these are from simulated 2D images on white backgrounds~\cite{xu2019disn}. \acp{DIAS} must operate in the challenging 3D medical setting, where datasets may be small and the contrast low. We formulate pose estimation as an agent-based \ac{MDP}, where the agent navigates to a pose matching the present anatomical structure. This has connections to some recent registration approaches~\cite{liao2017artificial,krebs2017robust,ma2017multimodal}. However, estimating both non-rigid and rigid poses presents an extremely large search space. To deal with this, 1) we introduce inverted episodic training and 2) we employ a deep realization of \ac{MSL}~\cite{zheng2014marginal,Zheng_2008} to incrementally learn pose parameters. The use of \ac{MSL} links \acp{DIAS} to a variety of traditional \ac{SSM} works pre-dating modern deep networks.

\begin{figure*}[t]
\centering
        \includegraphics[width=.72\linewidth]{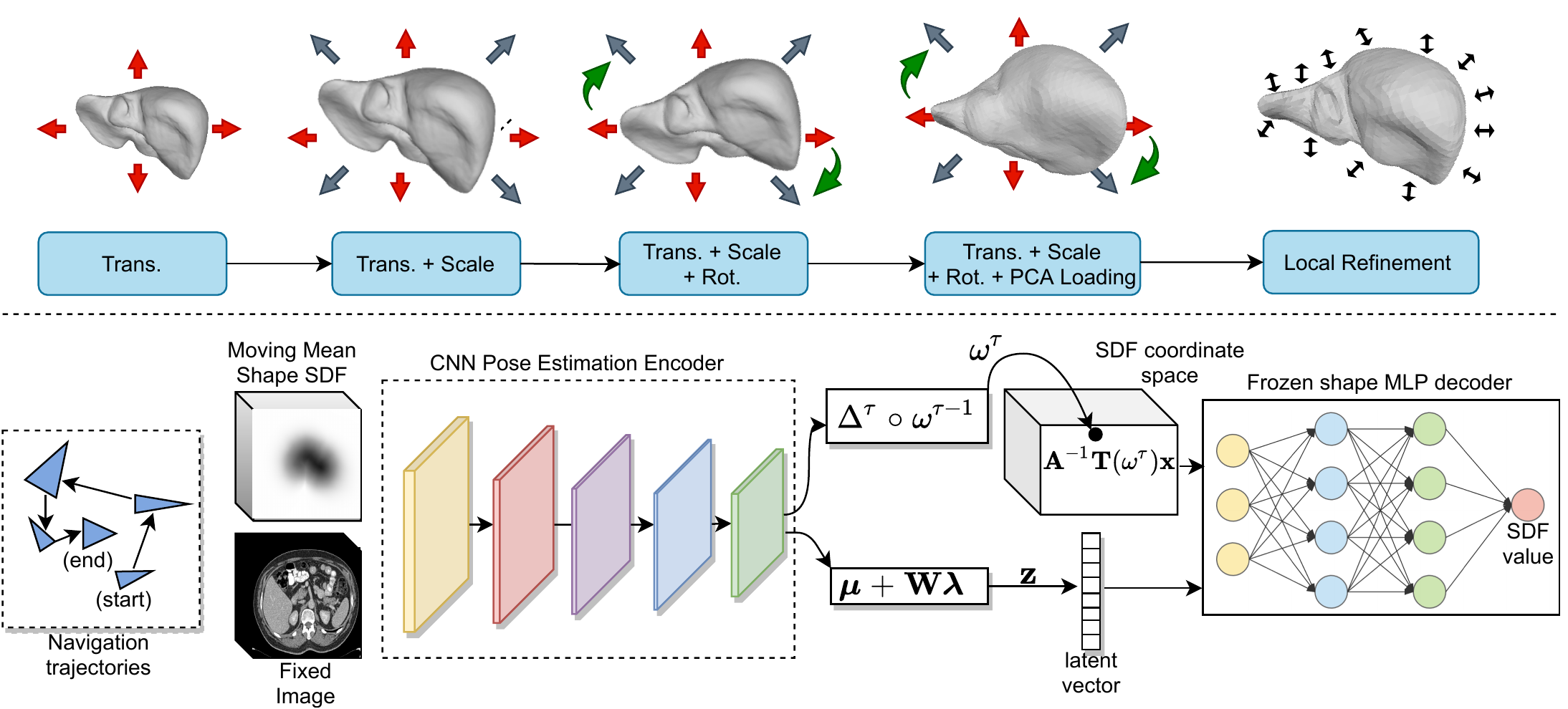}

   \caption{Overview of our deep implicit statistical shape modeling (\acs{DIAS}) framework. }
\label{fig:system} 
\end{figure*}
\label{sec:shape_model}

\section{Method}
Fig.~\ref{fig:system} illustrates the \ac{DIAS} framework. As the bottom panel demonstrates, a \ac{CNN} encoder predicts rigid and non-rigid poses, which, along with desired coordinates, are fed into a deep implcit \ac{MLP} shape decoder to output corresponding \ac{SDF} values. The encoder searches for the best pose using an \ac{MDP} combined with \ac{MSL} (top panel). We first outline the deep implicit shape model, discuss pose estimation, then describe the final local surface refinement.

\subsection{Implicit Shape Representation}

Deep implicit shapes~\cite{park2019deepsdf,chen_learning_2019,mescheder_occupancy_2019} are a recent and powerful implicit shape representation. We use \citet{park2019deepsdf}'s formulation to model organ shapes using an \ac{SDF}, which, given a coordinate, outputs the distance to a shape's surface:
\begin{align}
    \SDF(\mathbf{x})= s: x \in \mathbb{R}^3, \, s \in \mathbb{R} \mathrm{,} \label{eqn:sdf}
\end{align}
where $s$ is negative inside the shape and positive outside of it. The iso-surface of \eqref{eqn:sdf}, \ie, coordinates where it equals $0$, corresponds to the shape surface. The insight of deep implicit shapes is that given a set of coordinate/\ac{SDF} pairs \emph{in some canconical or normalized space}, $\tilde{\mathcal{X}} = \{\tilde{\mathbf{x}}_{i}$, $s_{i}\}$, a deep \ac{MLP}  can be trained to approximate a shape's SDF:
\begin{align}
f_{\theta_{S}}(\tilde{\mathbf{x}}) \approx \SDF(\tilde{\mathbf{x}}), \, \forall \tilde{\mathbf{x}} \in \Omega \mathrm{,} \label{eqn:deep_sdf_1}
\end{align}
where the tilde denotes canonical coordinates. Because it outputs smoothly varying values \eqref{eqn:deep_sdf_1} has no resolution limitations, and, unlike meshes, it does not rely on an explicit discretization scheme. In practice the resolution of any captured details are governed by the model capacity and the set of training samples within $\mathcal{X}_{i}$. While \eqref{eqn:deep_sdf_1} may describe a single shape, it would not describe the anatomical variation across a set of shapes. To do this, we follow \citet{park2019deepsdf}'s elegant approach of auto-decoding latent variables. We set of $K$ \ac{SDF} samples from the same organ, $\mathcal{S}=\{\tilde{\mathcal{X}}_{k}\}_{k=1}^{K}$, and we create a corresponding set of latent vectors, $\mathcal{Z}=\{\mathbf{z}_{k}\}_{k=1}^{K}$. The deep \ac{MLP} is modified to accept two inputs, $f_{\theta_{S}}(\tilde{\mathbf{x}}, \mathbf{z}_{k})$, conditioning the output by the latent vector to specify which particular \ac{SDF} is being modeled. We jointly optimize the network weights $\theta_{S}$ and latent vectors $\mathcal{Z}$ to produce the best \ac{SDF} approximations: 
\begin{align}
    \argmin_{\theta_{S}, \mathcal{Z}} \sum_{k}^{K}\left(\sum_{i=1}^{|\tilde{\mathcal{X}_{k}}|}  \mathcal{L}(f_{\theta_{S}}(\tilde{\mathbf{x}}_{i}, \mathbf{z}_{k}),s_{i}) + \dfrac{1}{\sigma^2}\|\mathbf{z}_{k} \|_2^2\right) \mathrm{,} \label{eqn:shape_loss}
\end{align}
where $\mathcal{L}$ is the L1 loss and the second term is a zero-mean Gaussian prior whose compactness is controlled by $\sigma^2$. 

The implicit shape model assumes shapes are specified in a canonical space. However, segmentation labels are usually given as 3D masks. To create canonical coordinate/\ac{SDF} training pairs, we first perform within-slice interpolation~\cite{Albu_2008} on masks to remove the most egregious of discretization artifacts. We then convert the masks to meshes using marching cubes~\cite{Lewiner_2003}, followed by a simplification algorithm~\cite{Quadric}. Each mesh is then rigidly aligned to an arbitrarily chosen anchor mesh using coherent point drift~\cite{Myronenko_2010}. Similar to~\citet{park2019deepsdf}, \ac{SDF} and coordinate values are randomly sampled from the mesh, with regions near the surface much more densely sampled. \ac{SDF} values are also scaled to fit within $[-1,1]$. Based on the anchor mesh, an affine matrix that maps between canonical and pixel coordinates can be constructed, $\mathbf{x}=\mathbf{A}\tilde{\mathbf{x}}$. More details can be found in the Supplementary.

\begin{figure}[t]
    \centering
    \small
    \renewcommand{\arraystretch}{0}
    \footnotesize
    \begin{tabular}{ccccc}
     \includegraphics[width=\renderwidth]{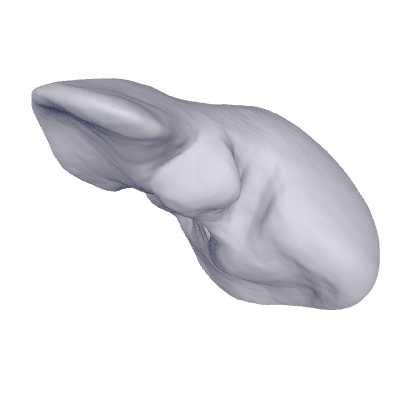} & \includegraphics[width=\renderwidth]{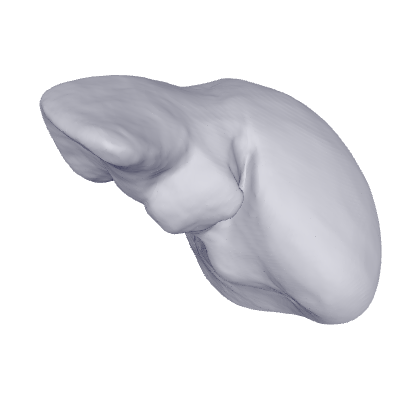} & \includegraphics[width=\renderwidth]{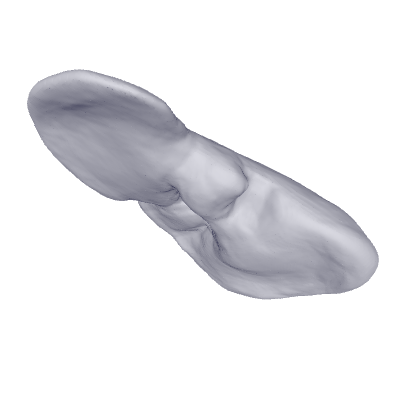} & \includegraphics[width=\renderwidth]{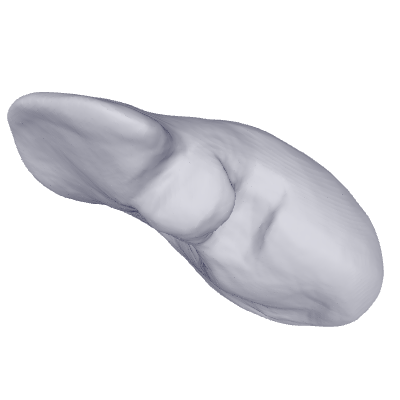} & \includegraphics[width=\renderwidth]{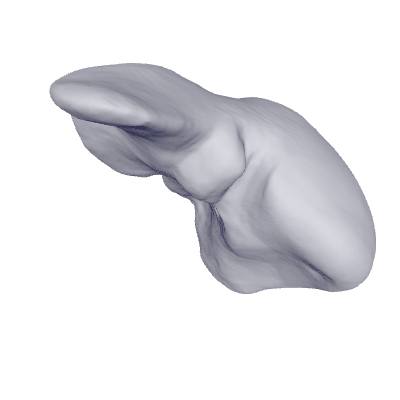} \\
     mean & $\lambda_1 = -0.5$ & $\lambda_1 = 0.5$ & $\lambda_2 = -0.5$ & $\lambda_2 = 0.5$ \\
     \includegraphics[width=\renderwidth]{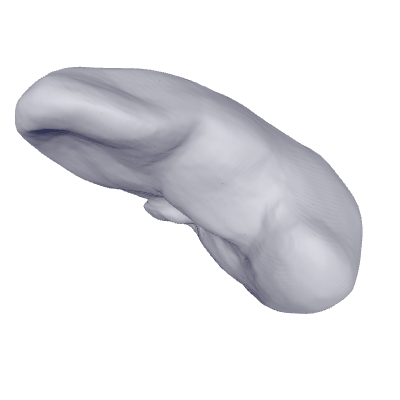} & \includegraphics[width=\renderwidth]{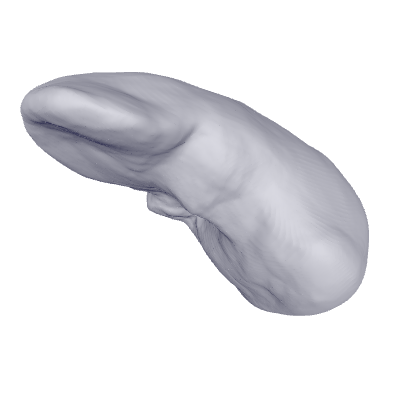} & \includegraphics[width=\renderwidth]{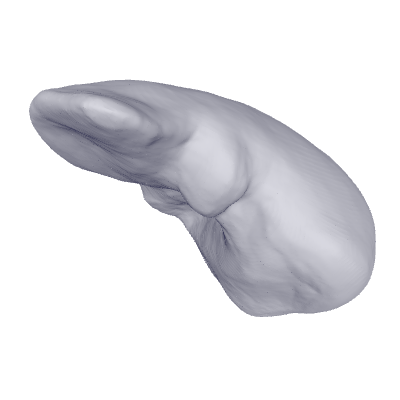} & \includegraphics[width=\renderwidth]{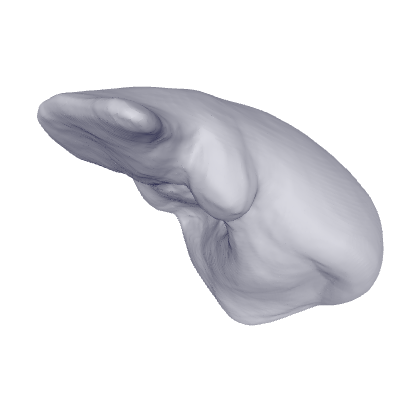} & \includegraphics[width=\renderwidth]{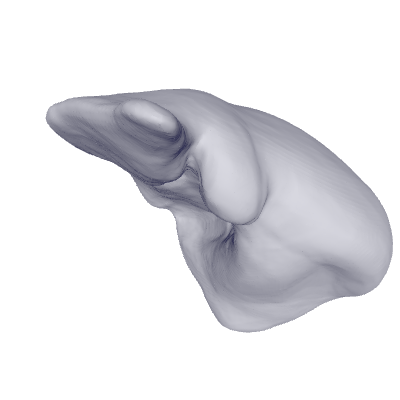} \\
     $\alpha=0$ & $\alpha=0.25$ & $\alpha=0.5$ & $\alpha=0.75$ & $\alpha=1$ \\
    \end{tabular}
    
    \renewcommand{\arraystretch}{1}
    \caption{\acs{DIAS} shape embedding space on \acs{MSD} liver dataset~\cite{Simpson_2019}. The top row shows the shapes generated from the mean latent vector and scaled version of the two first \acs{PCA} bases ($\lambda_1$ and $\lambda_2$). The bottom row renders an interpolation between two selected latent vectors: $\mathbf{z} = (1-\alpha)\mathbf{z}_{0} + \alpha\mathbf{z}_{1}$. }
    \label{fig:shape} 
\end{figure}

Once the shape decoder is trained, any latent vector can be inputted into $f_{\theta_{S}}(.,.)$ along with a set of coordinates to rasterize an \ac{SDF}, which can then be rendered by extracting the iso-boundary. As the top row of Fig.~\ref{fig:shape} and bottom panel of Fig.~\ref{fig:intro} demonstrate, the latent space provides a rich description of shape variations. The mean latent vector, $\boldsymbol\upmu$, produces an anatomically valid shape. A \ac{PCA} can capture meaningful variation, \eg{}, the first basis corresponds to stretching and flattening while the second controls the prominence of lobe protuberances. Interpolating between latent vectors produces reasonable shapes (bottom row of Fig.~\ref{fig:shape}).

\subsection{Pose Estimation}
\label{sec:pose}

The next major step is use the compact and rich \ac{DIAS} shape space to delineate an object boundary given an image, $I$. We assume a dataset of labelled images is available, allowing for the generation of coordinate/\ac{SDF} pairs: $\mathcal{D}=\{I_{k}, \mathcal{X}_{k}\}_{k=1}^{K_{\mathcal{D}}}$, where $\mathcal{X}_{k} = \{\mathbf{x}_{i}$, $s_{i}\}$ is specified using pixel coordinates. Note, we only assume a mask/\ac{SDF} is present, and do not require \emph{explicit} ground-truth rigid and non-rigid poses. We need to define: 1) the rigid-body transform from the canonical coordinates to the image space, and 2) the latent vector that specifies the shape variation. We denote the rigid-body transformation as $\mathbf{T}(\omega)$ with parameters $\omega=\{\mathbf{t}, \mathbf{s}, \mathbf{b}\}$, \ie{}, translation, anisotropic scale, and rotation, respectively, where $\mathbf{t}\in \mathbb{R}^3$, $\mathbf{s}\in \mathbb{R}^3$, and $\mathbf{b}\in \mathbb{R}^6$. Here we use \citet{zhou_continuity_2019}'s recent six-dimensional parameterization of the rotation matrix, where we actually predict deviations from identity, \ie{}, $\mathbf{I} + \mathbf{T}(\mathbf{b})$. We model shape variation using a truncated \ac{PCA} basis that only captures salient variations: $\mathbf{z}=\boldsymbol\upmu+\mathbf{W}\boldsymbol\uplambda$. Unlike explicit \acp{SSM}, the \ac{PCA} is performed on the latent space and does not require correspondences. We employ an encoder network, $g_{\theta_{E}}(I)$, to predict the rigid pose parameters, $\omega$, and the non-rigid \ac{PCA} loadings $\boldsymbol\uplambda$. The parameters predicted by $g_{\theta_{E}}(.)$ are fed into a frozen $f_{\theta_{S}}(.,.)$ to produce the object's \ac{SDF}:
\begin{align}
     \SDF(\mathbf{x}) &= f_{\theta_S}\left(\mathbf{A}^{-1}\mathbf{T}(\omega)\mathbf{x},\boldsymbol\upmu+\mathbf{W}\boldsymbol\uplambda \right), \label{eqn:formulation} \\
    \omega,  \boldsymbol\uplambda &= g_{\theta_E}(I) \textrm{.} \label{eqn:formulation_2}
\end{align}

While \eqref{eqn:formulation} and \eqref{eqn:formulation_2} could work in principle, directly predicting global pose parameters in one shot is highly sensitive to any errors and we were unable to ever reach convergence. We instead interpret the encoder $g_{\theta_E}(.)$ as an ``agent'' that, given an initial pose, $\omega^0$, generates samples along a trajectory by predicting corrections to the previous pose:
\begin{align}
    \Delta^\tau, \,\boldsymbol\uplambda^\tau &= g_{\theta_E}\left( I_{k}, \omega^{\tau - 1} \right) \mathrm{,} \label{eqn:encoder_trajectory_3} \\
    \omega^{\tau} &=  \Delta^\tau \circ \omega^{\tau - 1}, \quad \text{if } \tau>0  \mathrm{,}  \label{eqn:encoder_trajectory_4}
\end{align}
where $\circ$ denotes the composition of two rigid-body transforms and $\tau$ indicates the current step in the trajectory. An observation of $\omega^{\tau-1}$ is injected into the input of the encoder so that it is aware the previous pose to correct. To do this we rasterize  the \ac{SDF} corresponding to the mean shape, $\SDFmu$, \emph{once}. After every step it is rigidly transformed using $\omega^{\tau-1}$ and fed as a second input channel into $g_{\theta_E}(.,.)$. The agent-based formulation turns the challenging one-step pose estimation task into a simpler multi-step correction task. Note in \eqref{eqn:encoder_trajectory_3} we do not predict a trajectory for the \ac{PCA} loadings. Unlike rigid pose estimation, which can use the transformed $\SDFmu$, it not clear how to best inject a concept of \ac{PCA} state into the encoder without rasterizing a new \ac{SDF} after every step. Since this is prohibitively expensive, the \ac{PCA} loadings are directly estimated at each step. We break the search space down even further by first predicting rigid poses then predicting the \ac{PCA} loadings, as detailed below. 

\subsubsection{Rigid Pose Estimation}

We first train the encoder $g_{\theta_E}(.,.)$ to predict rigid poses. In training we generate samples along a trajectory of $\Tau$ steps, which is referred to as an episode. The encoder is trained by minimizing a loss calculated over the episodes generated on the training data:
\begin{align}
    \argmin_{\theta_E} & \sum_{k=1}^{K_{\mathcal{D}}}\sum_{i=1}^{|\mathcal{X}_{k}|}  \sum_{\tau=1}^{\Tau}\mathcal{L}(f_{\theta}\left(\mathbf{A}^{-1}\mathbf{T}(\omega^{\tau})\mathbf{x}_{i},\boldsymbol\upmu \right),s_{i} ) \mathrm{,} \label{eqn:rigid_loss}
\end{align}
where back-propagation is only executed on the encoder weights $\theta_E$ within each step $\tau$, and the dependence on $g_{\theta_E}(.)$ is implied through $\omega_{\tau}$. Note that \eqref{eqn:rigid_loss} uses the mean latent vector, $\boldsymbol\upmu$, to generate \ac{SDF} predictions and the $\,\boldsymbol\uplambda$ output in \eqref{eqn:encoder_trajectory_3} is ignored for now. This training process shares similarities to \ac{DRL}, particularly in its formulation of the prediction step as an \ac{MDP}. Unlike \ac{DRL}, and similar to \ac{MDP} registration tasks~\cite{liao2017artificial,krebs2017robust,ma2017multimodal}, there is no need for cumulative rewards because a meaningful loss can be directly calculated. At the start of training, the agent will not produce reliable  trajectories but, as the model strengthens, the playing out of an \emph{episode} of $\Tau$ steps for each training iteration will better sample meaningful states to learn from. To expose the agent to a greater set of states, we also inject random pose perturbations, $\eta$, after every step, modifying \eqref{eqn:encoder_trajectory_3} and \eqref{eqn:encoder_trajectory_4} to
\begin{align}
\Delta^\tau, \,\boldsymbol\uplambda^\tau &= g_{\theta_E}\left( I_{k}, \eta^{\tau} \circ \omega^{\tau - 1} \right) \mathrm{,} \label{eqn:encoder_trajectory_3_noise} \\
\omega^{\tau} &=  \Delta^\tau \circ \eta^{\tau} \circ \omega^{\tau - 1}, \quad \text{if } \tau>0  \mathrm{.}  \label{eqn:encoder_trajectory_4_noise}
\end{align}
A downside to episodic training is that each image is sampled consecutively $\Tau$ times, which can introduce instability and overfitting to the learning process. To avoid this we introduce \emph{inverted episodic training}, altering the loop order to make each episodic step play out as an outer loop:
\begin{align}
    \argmin_{\theta_E} & \sum_{\tau=1}^{\Tau}\sum_{k=1}^{K_{\mathcal{D}}}\sum_{i=1}^{|\mathcal{X}_{k}|}  \mathcal{L}(f_{\theta}\left(\mathbf{A}^{-1}\mathbf{T}(\omega^{\tau})\mathbf{x}_{i},\boldsymbol\upmu \right),s_{i} ) \mathrm{,} \label{eqn:encoder_trajectory_1_inverted} 
\end{align}
where $\omega^{\tau}$ is saved for each sample after each iteration. 


\noindent{\textbf{Marginal Space Learning}:}
The \ac{MDP} of \eqref{eqn:encoder_trajectory_1_inverted} provides an effective sampling strategy, but it  requires searching amongst all possible translation, scale, and rotation configurations, which is too large a search space. Indeed we were unable to ever reliably produce trajectories that converged. To solve this, \acp{DIAS} use a deep realization of \acf{MSL}~\cite{zheng2014marginal}. \ac{MSL} decomposes the search process into a chain of dependant estimates, focusing on one set while marginalizing out the others. In practice (Tab.~\ref{tab:tbl:MSL}) this means that we first limit the search space by training the encoder to only predict a translation trajectory, $\mathbf{t}$, with the random perturbrations also limited to only translation, \ie{}, $\eta_{t}$. 
\begin{table}
\small
    \centering
    \begin{tabular}{c|c|c|c}
         Stage & $\Delta$ & $\eta$ & $\omega^{0}$ \\
         \hline
         Trans. & $\mathbf{t}$ & $\eta_{t}$ & $\omega_{\mathcal{D}}$ \\
         Scale & $\{\mathbf{s},\,\mathbf{t}\}$ & $\eta_{s}\circ\eta'_{t}$ & $\{\omega_{t,k}^{\Tau}\}_{k=1}^{K_\mathcal{D}}$\\
         Rot. & $\{\mathbf{b},\,\mathbf{s},\,\mathbf{t}\}$ & $\eta_{r}\circ \eta'_{s}\circ\eta'_{t}$ &  $\{\omega_{s,k}^{\Tau}\}_{k=1}^{K_\mathcal{D}}$\\
         Non Rig. & $\{\mathbf{b},\,\mathbf{s},\,\mathbf{t}\}$ & $\eta'_{r}\circ \eta'_{s}\circ\eta'_{t}$ &  $\{\omega_{r,k}^{\Tau}\}_{k=1}^{K_\mathcal{D}}$
    \end{tabular}
    
    \caption{\Acs{MSL} schedule used in \acs{DIAS}.}
    \label{tab:tbl:MSL} 
\end{table}
The initial pose is the mean location in the training set, denoted $\omega_{\mathcal{D}}$. Once trained, the translation encoder weights and final poses, $\{\omega_{t,k}^{\Tau}\}_{k=1}^{K_\mathcal{D}}$, are used to initialize a scale encoder of identical architecture, but one that predicts  scale corrections, $\mathbf{s}$, in addition to finetuning the translation. Importantly, to focus the search space on scale, the random \emph{translation} perturbations are configured to be much smaller than before, which is represented by the prime modifier on $\eta'_{t}$. Finally, a rotation model is trained (while finetuning translation + scale with smaller perturbations). In inference, the rigid pose is estimated by successively applying the models of each stage, using the final pose of the previous step to initialize the next.

\subsubsection{Non-Rigid Pose Estimation}

Once a rigid pose is determined, anatomically plausible deformations can then be estimated. We initialize the weights and poses of the non-rigid encoder using the translation + scale + rotation rigid model, modifying \eqref{eqn:encoder_trajectory_1_inverted} to now incorporate the \ac{PCA} basis:
\begin{align}
    \argmin_{\theta_E} & \sum_{\tau=1}^{\Tau}\sum_{k=1}^{K_{\mathcal{D}}}\sum_{i=1}^{|\mathcal{X}_{k}|}  \mathcal{L}(f_{\theta}\left(\mathbf{A}^{-1}\mathbf{T}(\omega_{\tau})\mathbf{x}_{i},\boldsymbol\upmu+\mathbf{W}\boldsymbol\uplambda^\tau \right),s_{i} )  \nonumber \\
    &+ \dfrac{1}{\sigma^2}\|\boldsymbol\upmu+\mathbf{W}\boldsymbol\uplambda^\tau\|_2^2\mathrm{.} \label{eqn:encoder_trajectory_1_pca} 
\end{align}
As Table~\ref{tab:tbl:MSL} indicates, the random rigid perturbations are configured to be small in magnitude to confine the search space to primarily the \ac{PCA} loadings. 

\subsection{Surface Refinement}

Like classic \acp{SSM}~\cite{Heimann_2009}, non-rigid pose estimation  provides a robust and anatomically plausible prediction, but it may fail to capture very fine details. We execute local refinements using an \ac{FCN} model, $r=h_{\theta_R}(I, \SDFpca)$, that accepts a two-channel input comprising the 3D image and the rasterized \ac{SDF} after the non-rigid shape estimation. Its goal is to refine the $\SDFpca$ to better match the ground truth \ac{SDF}. To retain an implicit surface representation, we adapt portions of a recent deep level set loss~\cite{michalkiewicz_implicit_2019}:  
\begin{align}
    \mathcal{L}_{r} &= \sum_{\mathbf{x}\in\Omega_{b}}\left(\SDF(\mathbf{x})^2\cdot\delta_{\epsilon}\left(\SDFpca(\mathbf{x})+r(\mathbf{x})\right)\right)^{1/2} \nonumber \\
    &+ \lambda_{1} \sum_{\mathbf{x}\in\Omega_{b}}(\|\nabla (\SDFpca(\mathbf{x})+r(\mathbf{x}))\|-1)^2 \nonumber \\
    &+ \lambda_2\sum_{\mathbf{x}\in\Omega_{b}}|\max(0,r(\mathbf{x})-\rho)|  \mathrm{,} \label{eqn:refinement}
\end{align}
where $\SDF$ is the ground truth \ac{SDF} and $\delta_{\epsilon}$ is a differentiable approximation of the Dirac-delta function. The first term penalizes mismatches between the iso-boundaries of the refined \ac{SDF} ground-truth. The second term ensures a unit gradient everywhere, guaranteeing that it remains a proper \ac{SDF}. See \citet{michalkiewicz_implicit_2019} for more details. The third term ensures the refinement does not deviate too much from $\SDFpca$ beyond a margin, $\rho$, otherwise it is free to deviate without penalty. Following standard level set practices, we only produce refinements within a narrow band, $\Omega_{b}$, around the $\SDFpca$ iso-boundary, which is also represented in the loss of \eqref{eqn:refinement}. Finally, in addition to standard data augmentations to $I$, we also independently augment $\SDFpca$ with random rigid and non-rigid pose variations, enriching the model's training set. 


\section{Experiments}
\begin{figure}[t]
    \centering
    \setlength\tabcolsep{0pt}
    \footnotesize
    \begin{tabular}{ccc}
     \includegraphics[width=\stageswidth]{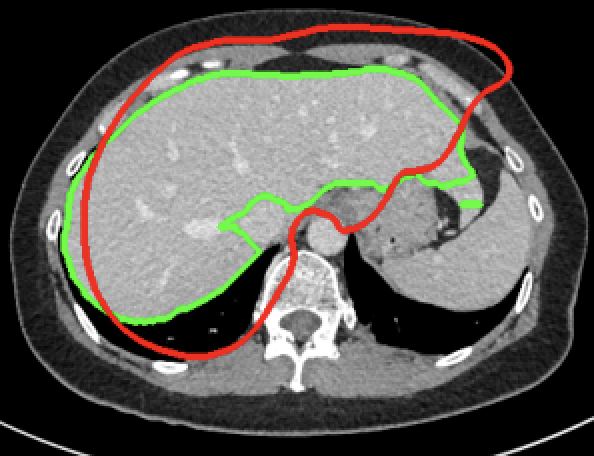} & \includegraphics[width=\stageswidth]{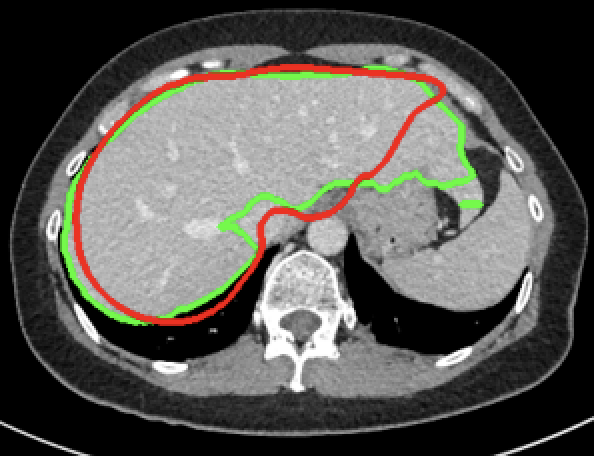} & \includegraphics[width=\stageswidth]{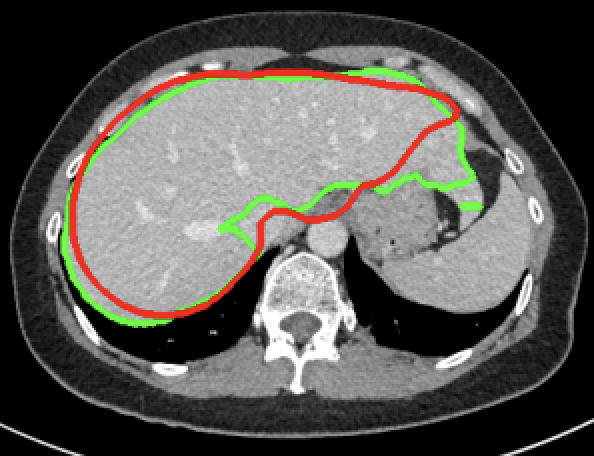} \\
     Translation &  Scale& Rotation \\
     \includegraphics[width=\stageswidth]{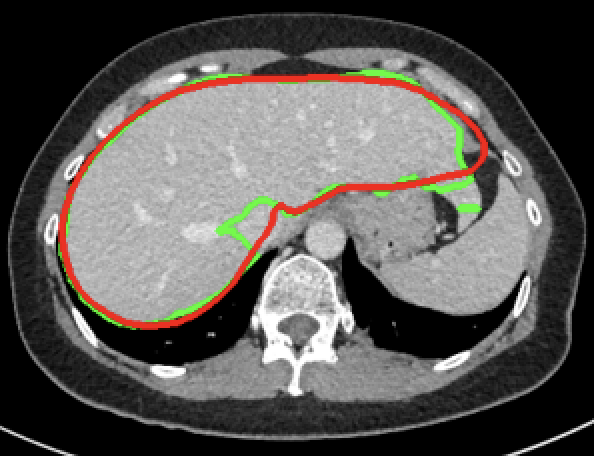} & \includegraphics[width=\stageswidth]{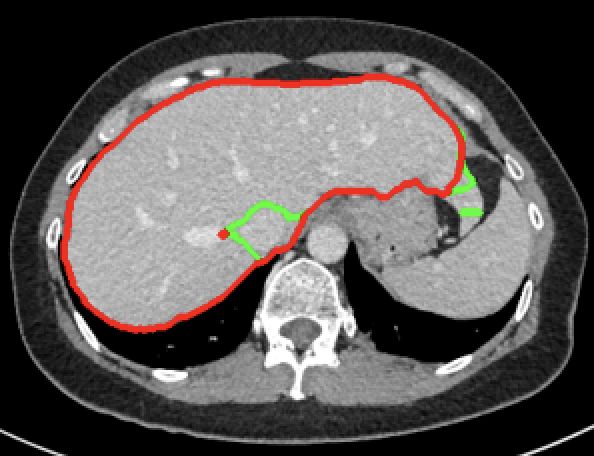} & 
     \includegraphics[width=\stageswidth]{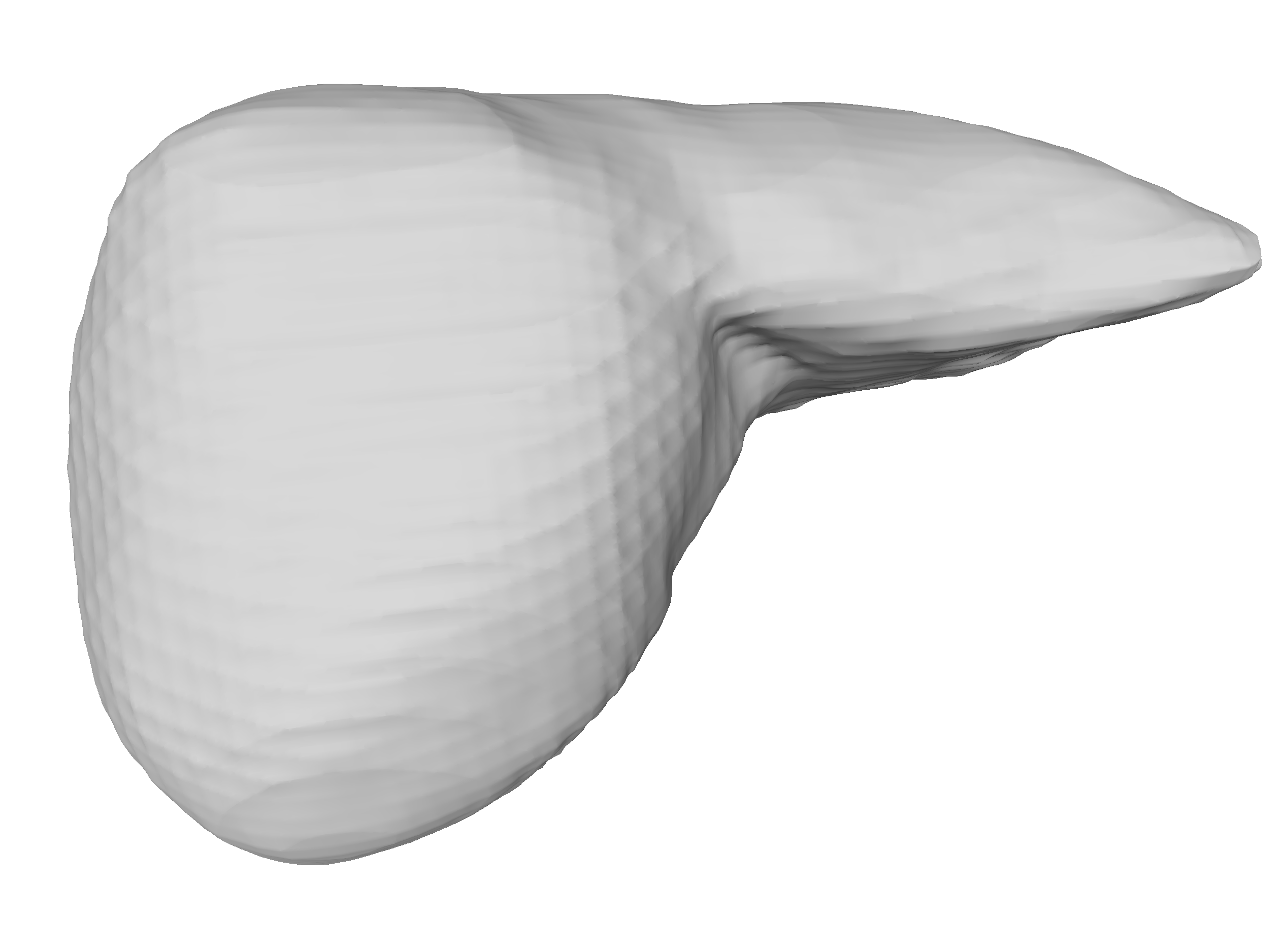}\\ 
       Non-rigid & Refinement & Surface
    \end{tabular}
    \renewcommand{\arraystretch}{1}
    \caption{The \ac{DIAS} pose estimation process. Red and green contours represent the prediction and ground truth, respectively.} 
    \label{fig:pose_stages}
\end{figure}
\noindent{\textbf{Liver Dataset}:} We focus on delineating pathological livers from venous-phase \acp{CT}. We use the size-$131$ training set of the \ac{MSD} liver dataset~\cite{Simpson_2019}, splitting it randomly into training, testing, and validation using proportions of $70\%$, $20\%$, and $10\%$, respectively. Tumor masks were merged into the liver to create a complete pathological liver mask. Intra-dataset results are important, but we take a further step to evaluate robustness and generalizability on unseen data from true and challenging clinical scenarios. To do this, we also evaluate on the external test set of \citet{Raju_2020}, which comprises $97$ venous-phase \acp{CT}. The dataset was sampled directly from a hospital archive with minimal curation and includes challenging scenarios not seen in the \ac{MSD} dataset, \ie{}, new lesion types and new co-morbitidies. Demographics are also different (Asian population vs. mostly Western population in \ac{MSD}). Thus, the external dataset helps reveal whether the \ac{DIAS} anatomical priors can provide robustness against unavoidable new and unseen clinical scenarios.
\begin{table}[t]
\centering
\small

\begin{tabular}{|l|l|l|l|l|l|l|l|}
\hline
\multicolumn{2}{|c|}{Model} & \multicolumn{2}{c|}{DSC (\%)} & \multicolumn{2}{c|}{ASSD (mm)} & \multicolumn{2}{c|}{HD (mm)} \\  \hline
\multicolumn{2}{|l|}{2D-PHNN}               & \multicolumn{2}{l|}{95.9 $\pm$ 2.7 }      &\multicolumn{2}{l|}{2.6 $\pm$ 1.2}  &\multicolumn{2}{l|}{35.7 $\pm$ 16.1} \\
\multicolumn{2}{|l|}{nnU-Net}            & \multicolumn{2}{l|}{96.4 $\pm$ 1.9}      & \multicolumn{2}{l|}{1.7 $\pm$ 0.9}   & \multicolumn{2}{l|}{29.1 $\pm$ 12.4}           \\ 
\multicolumn{2}{|l|}{H-DenseUNet}           & \multicolumn{2}{l|}{96.3 $\pm$ 2.1}      & \multicolumn{2}{l|}{1.9 $\pm$ 1.1} & \multicolumn{2}{l|}{30.4 $\pm$ 13.9}                 \\
\multicolumn{2}{|l|}{Adv. Shape Prior}           & \multicolumn{2}{l|}{96.0 $\pm$ 2.4}      & \multicolumn{2}{l|}{2.1 $\pm$ 1.3} & \multicolumn{2}{l|}{28.9 $\pm$ 12.8}                 \\
\multicolumn{2}{|l|}{\acs{DIAS} w/o refine}                  & \multicolumn{2}{l|}{96.1 $\pm$ 0.9}      & \multicolumn{2}{l|}{1.5 $\pm$ 0.7} & \multicolumn{2}{l|}{23.4 $\pm$ 12.2}                      \\ 
\multicolumn{2}{|l|}{\acs{DIAS} w refine}                  & \multicolumn{2}{l|}{\textbf{96.5 $\pm$ 0.7}}               & \multicolumn{2}{l|}{\textbf{1.1 $\pm$ 0.7}}                & \multicolumn{2}{l|}{\textbf{21.4 $\pm$ 11.8}}                    \\ \hline
\end{tabular}

\caption{Quantitative results on the MSD liver dataset.}
\label{tbl:MSD_LIVER}
\end{table}

\begin{figure}[t]
\centering
        \includegraphics[width=0.8\linewidth]{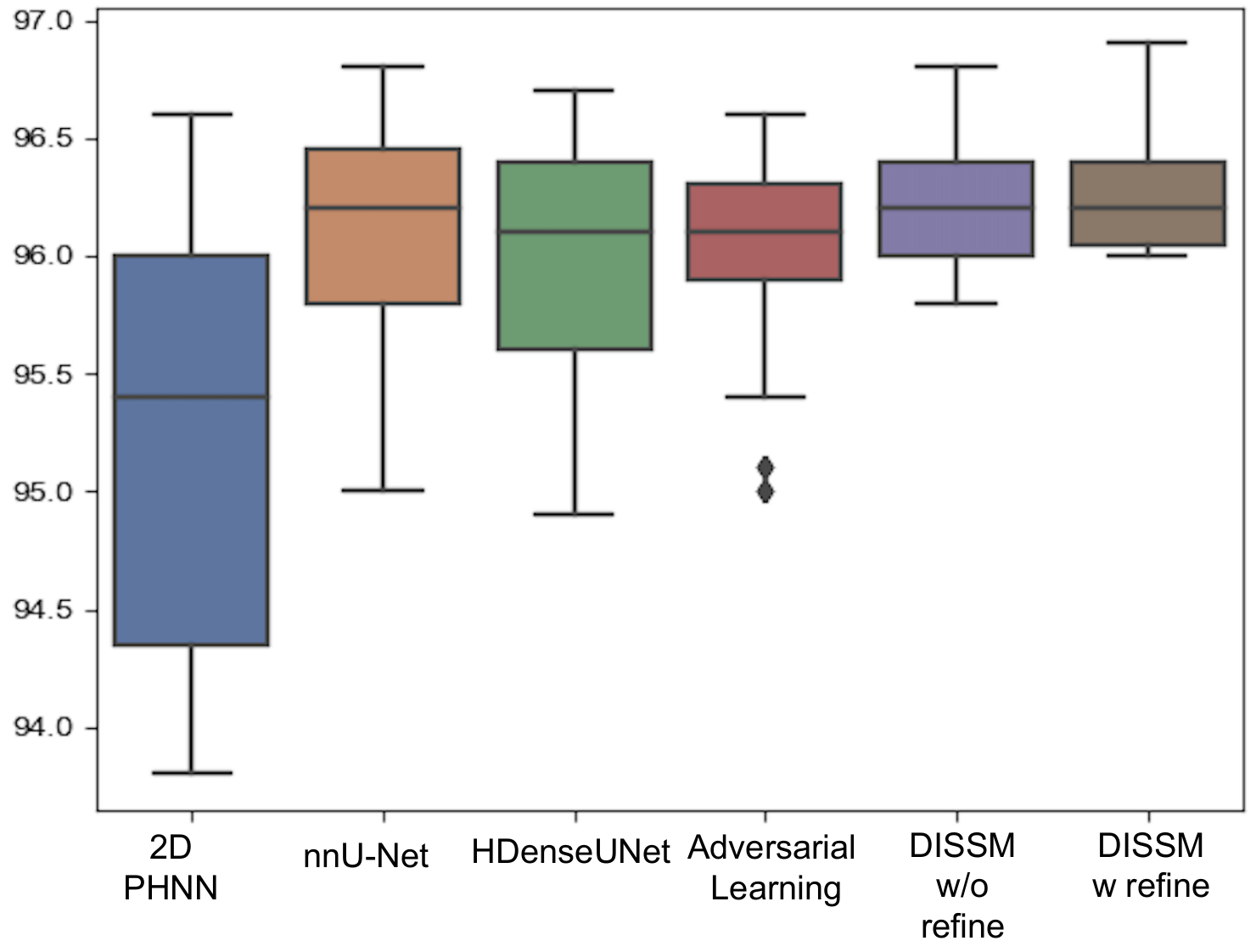}

\caption{Box-whisker plot of \acsp{DSC} on the MSD dataset.}

\label{fig:box_plot_msd}

\end{figure}
We compare \ac{DIAS} against very strong \ac{FCN} alternatives. 1) 2D P-HNN~\cite{Harrison_2017}, used as an \ac{FCN} backbone for a recent semi-supervised liver segmentation method~\cite{Raju_2020}. 2) HDenseUNet~\cite{Li_2018}, a leader of the LiTS liver segmentation challenge~\cite{Bilic_2019} that uses a cascade of 2D and 3D \acp{FCN}. 3) nnU-Net~\cite{isensee_nnu-net_2021}, the winner of the \ac{MSD}~\cite{Simpson_2019} and KiTS~\cite{heller2020state} challenges. We use its dual model 3D cascade setting, which performed best on the liver task of the \ac{MSD} challenge. 4) nnU-Net augmented with an adversarial anatomical prior that follows \citet{yang_automatic_2017}. For its discriminator, we use \citet{Raju_2020}'s more modern version, which has already proven effective on the clinical liver dataset. For all, we use their published implementations, including recommended resolutions, pre- and post-processing, and data augmentations. For quantitative comparisons we threshold the \ac{DIAS} \ac{SDF} surface to produce a mask and measure the \ac{DSC}, \ac{ASSD}, and \ac{HD} scores against the original masks. \emph{Note, this is a disadvantageous setup for \ac{DIAS}, as the original masks suffer from stair-like discretization effects, which \ac{DIAS} aims to rectify.}

\noindent{\textbf{Larynx Dataset}:} We also perform supplementary validation on larynx segmentation from \ac{CT}, a critical \ac{OAR} for head and neck cancer radiotherapy~\cite{brouwer2015ct}. This task is highly challenging due to very low contrast boundaries in \ac{CT}. We compare against the SOARS method and dataset of \citet{guo_organ_2020} ($142$ \acp{CT}), who reported a recent and computationally intensive approach that relies on stratified learning and neural architecture search. However, SOARS is also designed to segment other \acp{OAR} at the same time, so results are not apples-to-apples. For this reason, we additionally compare against nnU-Net~\cite{isensee_nnu-net_2021}, trained only to segment the larynx. More details on this dataset can be found in the supplementary.

\noindent{\textbf{Implementation Details}:} We kept settings as similar as possible for the two datasets. The shape decoder structure and hyperparameters follow that of \citet{park2019deepsdf} and we use a size $256$ latent variable. For the pose encoders, $g_{\theta_E}(.,.)$,  use the 3D encoder of a 3D U-Net~\cite{Cicek_2016}, with $4$ downsampling layers and global averaging pooling to produce $\Delta^\tau$ and $\boldsymbol\uplambda$. For the liver we estimate the first $28$ \ac{PCA} components ($72\%$ of the variance). Larynx shape variations are more constrained, so we estimate the first $12$ components ($95\%$ variance). The number of inverted episodic steps, $\Tau$, for training the translation, scale, rotation, and non-rigid encoders was $7$, $15$, $15$, and $15$, respectively. The translation encoder was trained on a coarsely sampled volume. After it converged, we cropped volumes to encompass the maximum organ size and trained the remaining pose encoders on higher resolution volumes. Finally, we use a patch-based 3D U-Net~\cite{Cicek_2016} as the local surface refinement model, $h_{\theta_{R}}(.,.)$. Full training details, including the schedule of random pose perturbations and a complete listing of all hyperparameters, can be found in the supplementary.




\noindent{\textbf{Liver Results}:} \ac{DIAS} consumes $12-13$s to fully delineate a 3D volume. Fig.~\ref{fig:pose_stages} overlays the \ac{DIAS} results on top of a \ac{CT} scan after each stage. As can be seen, each stage progressively improves the result.  After the non-rigid \ac{PCA} loading, the delineation quality is already quite high, capturing a lobe curve not represented by the mean shape. The local refinement improves results even further by capturing fine-grained boundary curvatures. An ablation study can be found in the supplementary.

Table~\ref{tbl:MSD_LIVER} outlines the performance on the \ac{MSD} data. As can be seen, all models perform quite well, but \ac{DIAS}  exhibits less variability in \ac{DSC} and \ac{ASSD}, indicating better robustness. This is cogently illustrated by the \ac{HD} numbers, \ie{}, the worst-case distances for each volume. \ac{DIAS} dramatically improves the \ac{HD} numbers by roughly $26\%$ to $40\%$, resulting in much more reliable delineations. This robustness can be best seen by the box and whisker \ac{DSC} plot of Fig.~\ref{fig:box_plot_msd}, which shows \ac{DIAS} posting better worst-case and third-quartile performance. The visual impact of these improvements can be seen in the first column of Fig.~\ref{fig:qualitative}.


\begin{figure}[t]
\centering
\setlength\tabcolsep{0pt}
     \renewcommand{\arraystretch}{0}
\begin{tabular}{cccc}
     
      \includegraphics[width=\qualwidth, height=\qualheight]{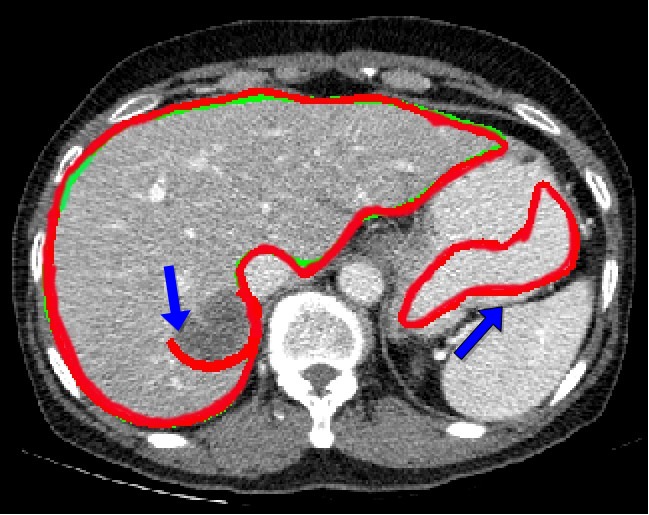} & \includegraphics[width=\qualwidth, height=\qualheight]{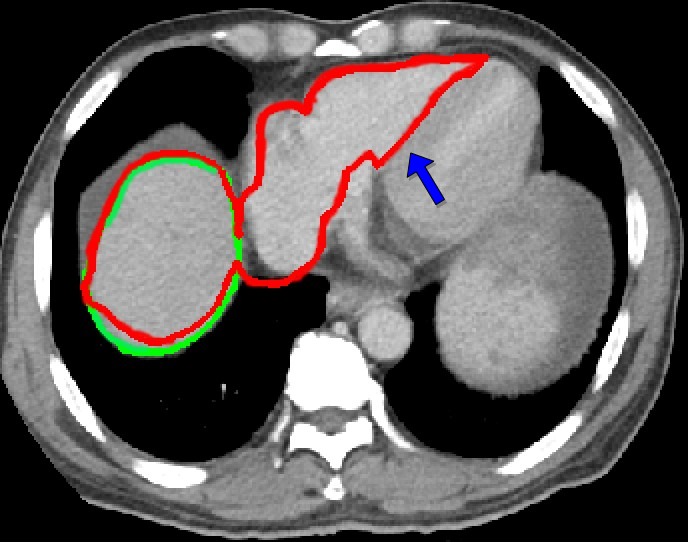} & \includegraphics[width=\qualwidth, height=\qualheight]{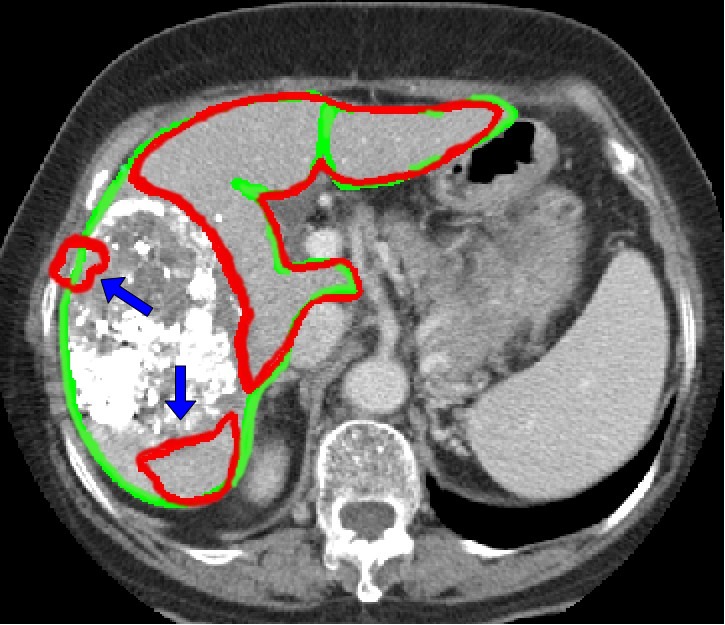} &
\includegraphics[width=\qualwidth, height=\qualheight]{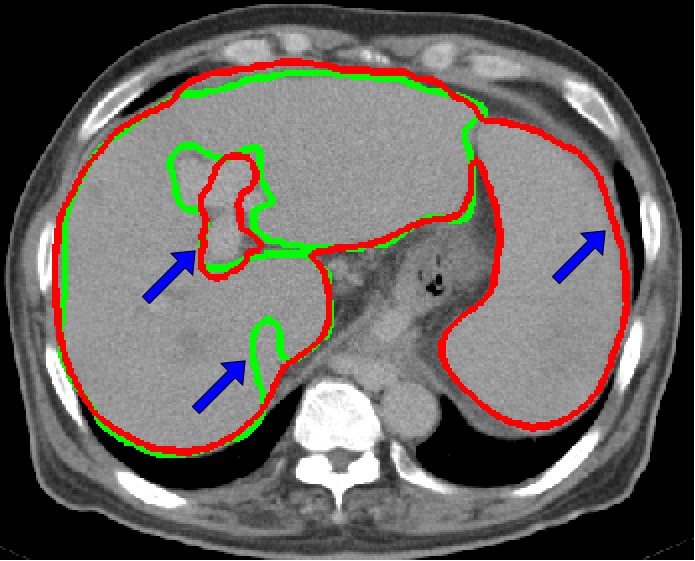} \\
 \includegraphics[width=\qualwidth, height=\qualheight]{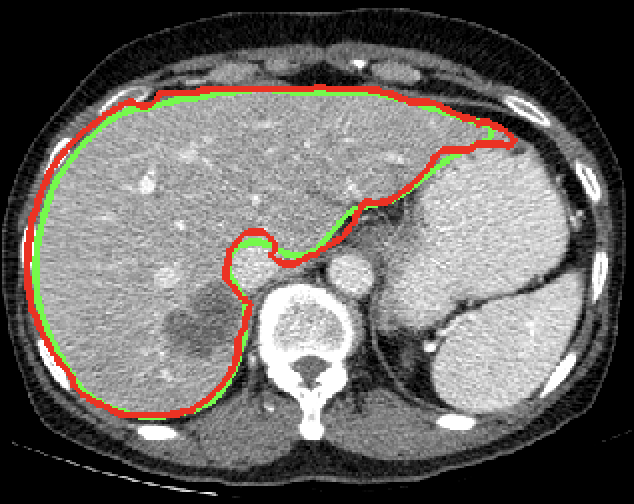} & \includegraphics[width=\qualwidth, height=\qualheight]{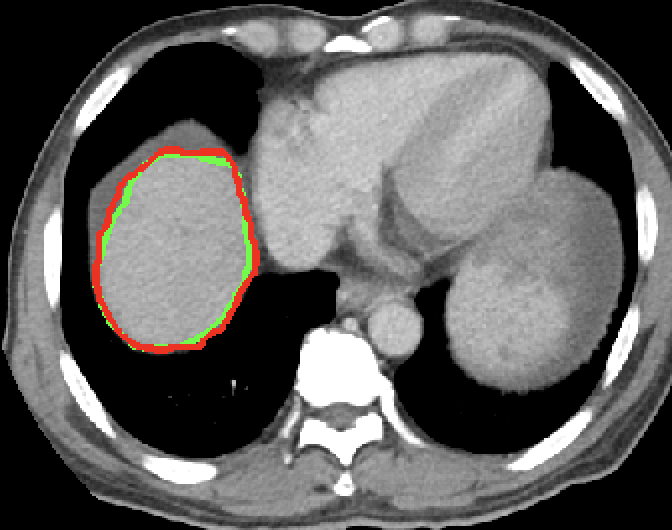} & \includegraphics[width=\qualwidth, height=\qualheight]{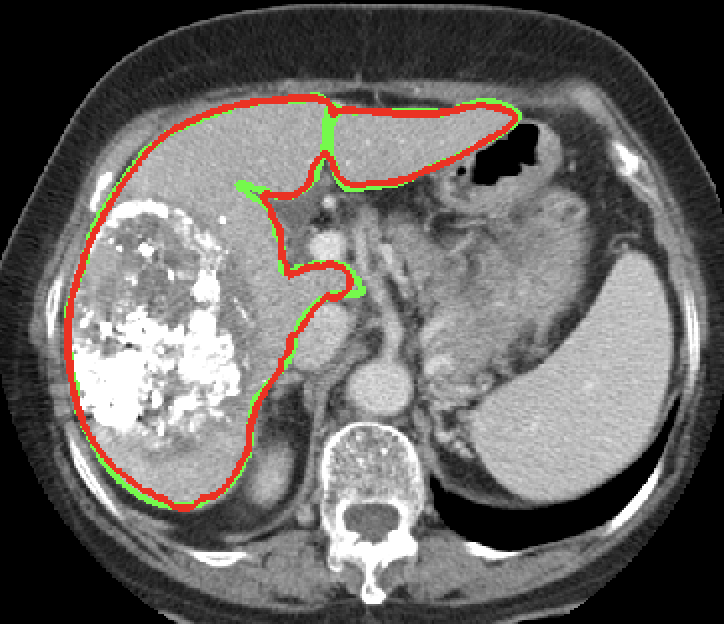} &
\includegraphics[width=\qualwidth, height=\qualheight]{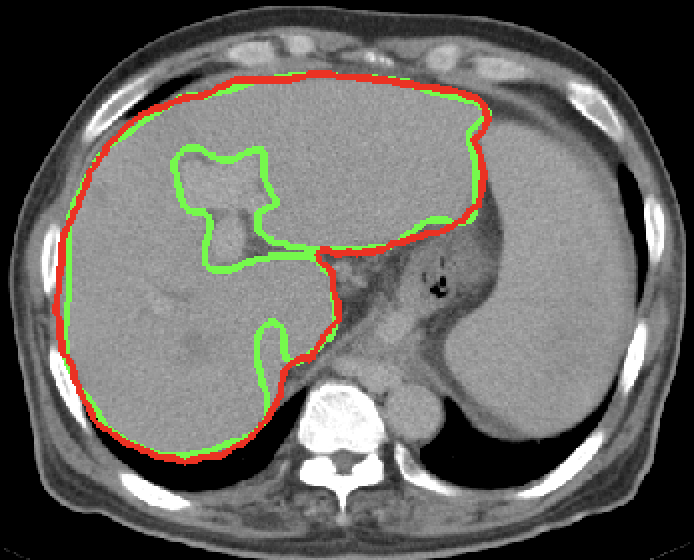}
\end{tabular}
\renewcommand{\arraystretch}{1}
\caption{Qualitative comparison of \acs{DIAS} (second row) versus nnU-Net (first row). Red and green contours represent the prediction and ground truth delineations, respectively. The first and second-to-fourth columns are drawn from the \ac{MSD} and clinical dataset, respectively. Fourth row is worst-case performance for \ac{DIAS}. Blue arrows in the top row shows nnU\_Net mispredictions. Results from other methods can be found in the Supplemental} 
    \label{fig:qualitative}
\end{figure}


\begin{table}[t]
\small
\centering
\begin{tabular}{|l|l|l|l|l|l|l|l|}
\hline

\multicolumn{2}{|c|}{Model} & \multicolumn{2}{c|}{DSC (\%)} & \multicolumn{2}{c|}{ASSD (mm)} & \multicolumn{2}{c|}{HD (mm)} \\  \hline
\multicolumn{2}{|l|}{2D-PHNN}               & \multicolumn{2}{l|}{90.1 $\pm$ 5.1}      &\multicolumn{2}{l|}{3.9 $\pm$ 1.4}   &\multicolumn{2}{l|}{46.3 $\pm$ 21.1} \\
\multicolumn{2}{|l|}{nnU-Net}            & \multicolumn{2}{l|}{92.4 $\pm$ 3.3}      & \multicolumn{2}{l|}{3.6 $\pm$ 1.1}    &\multicolumn{2}{l|}{34.1 $\pm$ 17.3}            \\
\multicolumn{2}{|l|}{HDenseUNet}           & \multicolumn{2}{l|}{92.1 $\pm$ 3.7}      & \multicolumn{2}{l|}{3.3 $\pm$ 1.3}   &\multicolumn{2}{l|}{36.2 $\pm$ 16.7}        \\
\multicolumn{2}{|l|}{Adv. Shape Prior}           & \multicolumn{2}{l|}{93.8 $\pm$ 1.7}      & \multicolumn{2}{l|}{3.1 $\pm$ 1.1} & \multicolumn{2}{l|}{31.7 $\pm$ 14.3}  \\
\multicolumn{2}{|l|}{\acs{DIAS} w/o refine}                  & \multicolumn{2}{l|}{95.7 $\pm$ 1.8}      & \multicolumn{2}{l|}{2.6 $\pm$ 1.1}     &\multicolumn{2}{l|}{24.7 $\pm$ 12.6}                 \\ 
\multicolumn{2}{|l|}{\acs{DIAS} w refine}                 & \multicolumn{2}{l|}{\textbf{95.9 $\pm$ 1.6}}      & \multicolumn{2}{l|}{\textbf{2.3 $\pm$ 0.9}}    &\multicolumn{2}{l|}{\textbf{21.8 $\pm$ 12.1}}                 \\ \hline
\end{tabular}

\caption{Cross-dataset results on the clinical liver dataset. }
\label{tbl:CGMH_LIVER} 
\end{table}

\begin{figure}[t]
\centering
        \includegraphics[width=.8\linewidth]{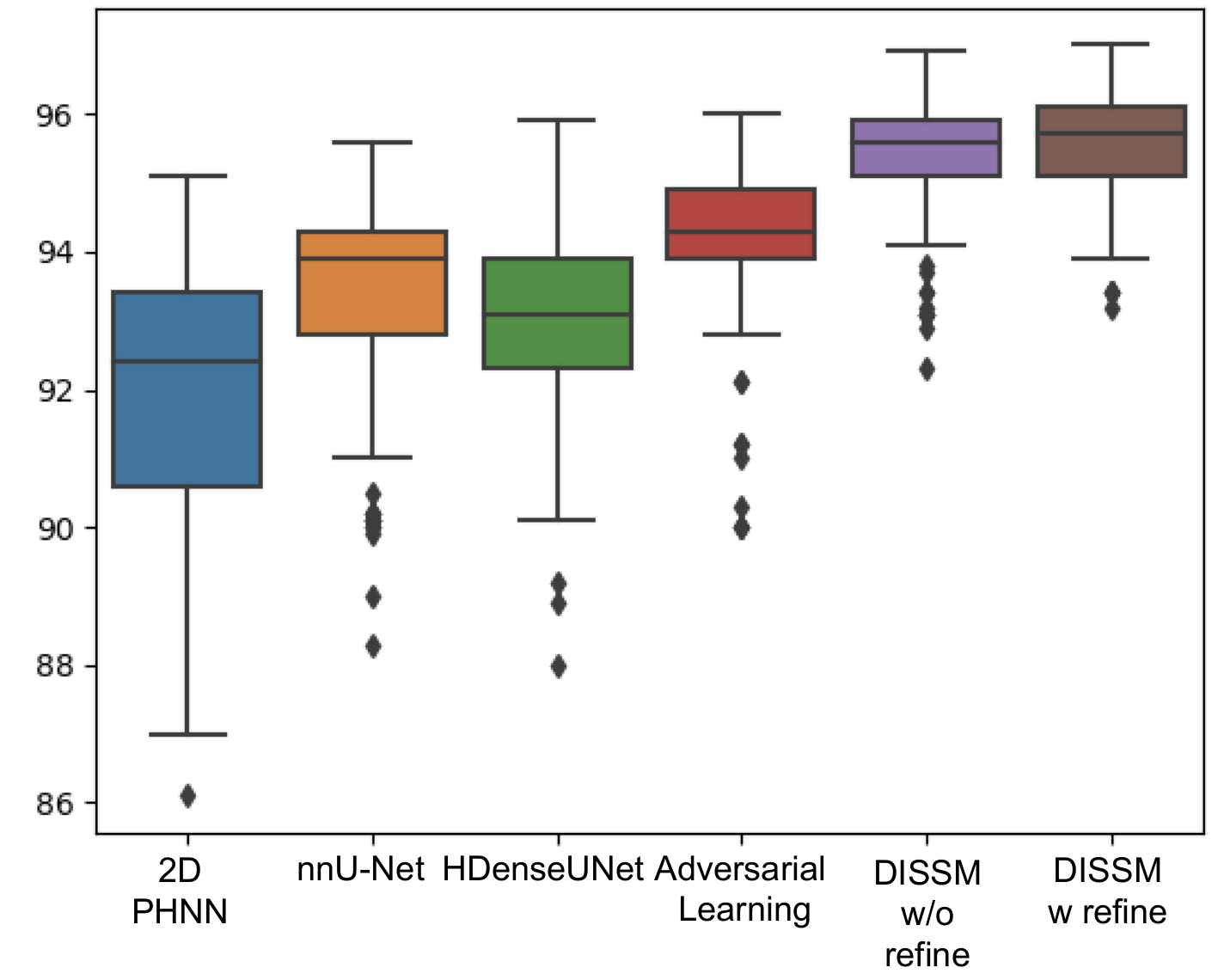}

\caption{Box-whisker plot of \acp{DSC} on the clinical dataset.}

\label{fig:box_plot_cgmh} 

\end{figure}

While the above demonstrate that \ac{DIAS} can provide superior intra-dataset robustness, the clinical results are even more telling. As Table~\ref{tbl:CGMH_LIVER} highlights, the competitor model performances drop drastically on the clinical dataset, underscoring the difficulty of operating when morbidities, scanners, patient populations, and practices can vary in unanticipated ways. In contrast, \ac{DIAS}'s performance is much more stable, still posting very good numbers. Compared to the competitors, \ac{DIAS} boosts the mean \ac{DSC} score by $2.1$ to $5.8\%$ and reduces the \ac{HD} by $31$ to $53\%$. As the box and whisker plot of Fig.~\ref{fig:box_plot_cgmh} shows, \ac{DIAS} also provides much better worst-case performance and smaller spread, even when compared against the adversarial prior.   
The best \ac{FCN}-only competitor, \ie{}, nnU-Net posts a worst-case \ac{DSC} performance of $88.1\%$, whereas \ac{DIAS}'s is a much better $93.2\%$. The second to fourth columns of Fig.~\ref{fig:qualitative} illustrate challenging clinical examples, where nnU-Net leaked into the cardiac region, failed to segment a treated lesion, and failed to handle a patient with splenomegaly. Note, the fourth row represents the worst-case result for \ac{DIAS}. Given that clinical deployments are the ultimate end goal of delineation solutions, these results offer convincing demonstration of the benefits of imposing strong and descriptive anatomical priors for segmentation. 

\begin{table}[t]
\small
\centering
\begin{tabular}{|l|l|l|l|l|l|l|l|}
\hline

\multicolumn{2}{|c|}{Model} & \multicolumn{2}{c|}{DSC (\%)} & 
\multicolumn{2}{c|}{HD (mm)} \\  \hline
\multicolumn{2}{|l|}{SOARS}               & \multicolumn{2}{l|}{56.7 $\pm$ 17.1}       &\multicolumn{2}{l|}{9.0 $\pm$ 7.1} \\
\multicolumn{2}{|l|}{nnU-Net}            & \multicolumn{2}{l|}{58.6 $\pm$ 14.7}     &\multicolumn{2}{l|}{8.6 $\pm$ 4.4}            \\
\multicolumn{2}{|l|}{\acs{DIAS} w/o refine}                  & \multicolumn{2}{l|}{59.4 $\pm$ 9.3}        &
\multicolumn{2}{l|}{ 7.8 $\pm$ 5.3}                 \\ 
\multicolumn{2}{|l|}{\acs{DIAS} w refine}                 & \multicolumn{2}{l|}{\textbf{60.9 $\pm$ 5.9}}        &\multicolumn{2}{l|}{\textbf{7.1 $\pm$ 5.7}}                 \\ \hline
\end{tabular}

\caption{Larynx dataset results. }
\label{tbl:larynx} 
\end{table}

\noindent{\textbf{Larynx Results}:} Table~\ref{tbl:larynx} outlines mean \ac{DSC} and \ac{HD} scores on the larynx dataset, which is all that SOARS reported~\cite{guo_organ_2020}. As can be seen, \ac{DIAS} outperforms the competitors, posting a mean \ac{DSC} of $60.9\%$, which is roughly $2\%$ better than the next best result (nnU-Net). More notably, the  standard deviation is significantly reduced (from $14.7$ to $5.9$), indicating that \ac{DIAS} is much more reliable. These robustness benefits on a delineation task with challenges distinct from pathological livers, \ie{}, low contrast boundaries, provide  further evidence of the value of the \ac{DIAS} \ac{SSM}. Qualitative examples can be found in the supplemental.

\section{Conclusion}

\Acfp{DIAS} use a deep implicit model to construct statistical and correspondence-free anatomical priors and directly outputs high-quality surfaces, rather than voxelized masks. For pose estimation, \ac{DIAS} proposes a robust \ac{MDP} that incorporates \ac{MSL} and inverted episodic training. \ac{DIAS} is the first to integrate a true \ac{SSM} with deep learning technology. Cross dataset evaluations on pathological liver segmentation demonstrate that \ac{DIAS} outperforms leading \acp{FCN}, \eg{}, nnU-Net~\cite{isensee_nnu-net_2021}, improving the mean and worst-case \ac{DSC} by $3.5\%$  and $5.1\%$, respectively. Supplemental validation on a challenging larynx dataset further confirmed the value of \ac{DIAS}. While continued maturation is necessary, \ac{DIAS} represents a new and promising approach to 3D medical imaging delineation.

\bibliography{egbib}

\appendix
\section{SDF Decoder Implementation}

\subsection{Sampling SDF Coordinate/Value Pairs}
\label{sec:sampling}
As mentioned in the main body, to train the shape decoder we require coordinate/\ac{SDF} value pairs in a canonical space. Since segmentation labels are almost always provided as voxelized masks, we must first remove the worst of the discretization effects. Because high inter-slice distances are a common issue, we first perform within-slice interpolation~\cite{Albu_2008} to a common $1$mm inter-slice resolution. We then apply Gaussian smoothing using the PyMCubes~\cite{PyMCubes} package to the masks, followed by marching cubes~\cite{Lewiner_2003} to produce a good quality mesh. Since this produces an unnecessarily dense set of vertices, we reduce their number using Quadric smoothing~\cite{Quadric}. We do this two times to  produce two versions: one reduced by a factor of $10$ and one by a factor of $100$, constructing a simplified and extremely simplified version of each mesh, respectively. The next step is to rigidly align each mesh to an arbitrarily chosen anchor mesh (the first mesh in the training list) using coherent point drift~\cite{Myronenko_2010,PyCPD}. We use the extremely simplified mesh versions since it is much more computationally efficient. The computed rigid alignment parameters from the ``extremely simplified'' meshes are then use to align the ``simplified'' meshes. 

We now have high-quality and aligned meshes, which are all scaled to fit in the unit sphere to create the canonical meshes. To sample \ac{SDF} values from the meshes, we follow Park \etal{}'s~\cite{park2019deepsdf} approach of randomly sampling within the unit sphere, with regions near the surface much more densely sampled. We use the mesh-to-sdf package~\cite{mesh2sdf}, which recreates the sampling scheme of Park \etal{}~\cite{park2019deepsdf}, but we modify its settings. First, we sample a greater proportion of uniformly distributed points ($20\%$ of coordinate/\ac{SDF} value pairs are sampled uniformly across the uniform sphere). The other $80\%$ are sampled closer to the boundary by randomly selecting boundary points and jittering them randomly. Unlike Park \etal{}~\cite{park2019deepsdf}, we use higher magnitude jitters: either $0.1$ or $0.01$ (with equal probability) in normalized space. This sampling scheme allowed our shape decoder to converge much better than when using Park \etal's{} settings. We sample $1$ million coordinate/\ac{SDF} value pairs per shape to train the \ac{SDF} decoder.

\subsection{Training}

For the most part, we follow Park \etal{}'s settings and hyper-parameters to train the shape decoder \ac{MLP}. Namely, the \ac{MLP} structure is a simple $8$ layer \ac{MLP} with $512$ channels in each layer and dropout with $20\%$ probability after each layer. The latent code is injected at the fourth layer. Weight normalization~\cite{Salimans_2016} is used after every layer as well. At every iteration and for every shape, $30\,000$ coordinate/\ac{SDF} value pairs are randomly sampled from the list of $1$ million to compute the loss. Like Park \etal{}, an equal proportion of positive and negative \ac{SDF} values are sampled. Unlike Park \etal{}~\cite{park2019deepsdf}, we do not use a clamped loss, since the specific \ac{SDF} values far away from the boundary are still useful for fitting pose. Table~\ref{tab:decoder} lists the hyper-parameters, most of which follow Park \etal{}~\cite{park2019deepsdf}. Note, after $1000$ epochs, the learning rates were reduced by a factor of $10$. 

\begin{table}
    \centering
    \begin{tabular}{|c|c|}
    \hline
       Parameter & Value \\ \hline
       $\sigma$  & $100$  \\
        Batch Size & $6$ \\
        \ac{SDF} Samples/Iteration & $30000$ \\
        $\mathcal{Z}$ Learning Rate & $0.001$ \\
        $\theta_{S}$ Learning Rate & $0.0005$ \\
        Epochs & $2000$ \\
        Latent Size & $256$ \\
        Dropout Probability & $0.2$ \\
        Optimizer & Adam~\cite{Kingma_2015}\\
         Weight decay & 0.0001 \\
        \hline
    \end{tabular}
    \caption{Hyper-parameters for training the shape decoder}
    \label{tab:decoder}
\end{table}

\section{Pose Encoder Implementation}

\subsection{Sampling SDF Coordinate/Value Pairs}

To train the pose encoders, coordinate/\ac{SDF} value pairs need to be sampled in the image space, \ie{}, not the canonical space. To do this, we first generate an \ac{SDF} from the original masks~\cite{Maurer_2003}. Like for training the \ac{SDF} decoder, we more densely sample closer to the shape boundary. We sample all coordinate/\ac{SDF} value pairs from regions within $13$ mm of the boundary (which corresponds to $10\%$ of the canonical space after normalization). We divide these up into positive and negative \ac{SDF} samples, and for each we add $10\%$ more coordinate/\ac{SDF} value pairs that are sampled uniformly across positive and negative regions of the entire \ac{SDF} volume, respectively. 

\subsection{Training}
We use the encoder structure of Lee \etal{}'s~\cite{lee2017superhuman} residual symmetric U-Net architecture to estimate the rigid and non-rigid pose parameters. The encoder structure has $4$ residual blocks with max pooling after each block. We perform global average pooling after the final layer of the final residual block, resulting in a size-$512$ feature vector. We train all pose encoders with an initial learning rate $0.001$ and with the Adam optimizer~\cite{Kingma_2015}, which is reduced by a factor of $10$ after each validation plateau. We truncate the image intensity values of all scans to the range of $[-160, 240]$ \ac{HU} to remove the irrelevant details. Table \ref{tab:encoder} lists other hyper parameters for training the \ac{DIAS} pose encoders.

\begin{table}
    \centering
    \begin{tabular}{|c|c|}
    \hline
       Parameter & Value \\ \hline
        Batch Size (Translation) & $8$ \\
        Batch Size (Scale, Rotation, Non-Rigid) & $12$ \\
        Epochs (Translation) & $65$ \\
        Epochs (Scale) & $80$ \\
        Epochs (Rotation) & $80$ \\
        Epochs (non-rigid) & $120$ \\
        \ac{SDF} Samples/Step & $150000$ \\
        $\sigma$ (Non-Rigid)  & $100$ \\
        Model Learning Rate & $0.001$ \\
        Weight Decay & $0.0001$ \\
        $\Tau$ (Translation) & $7$ \\
        $\Tau$ (Scale, Rotation, Non-Rigid) & $15$ \\
        \hline
    \end{tabular}
    \caption{Hyper-parameters for training \ac{DIAS}}
    \label{tab:encoder}
\end{table}

Following \ac{MSL} procedures, we first train a translation encoder to predict the translation needed for the mean shape to fit the volume. To estimate translation pose, the model needs to see the entire volume so that the necessary adjustments can be made to move closer to the ground truth. Due to limited computational resources, we resample the volume to $4 \times 4 \times 4$ mm resolution and symmetrically pad across the dimensions to create a unified volume size of physical dimension $512 \times 512 \times 808$ mm, which is large enough to cover all volumes in the \ac{MSD} training set.  To allow \ac{DIAS} to learn to provide pose corrections from a variety of translation trajectories, we perturb the prior pose at every episodic step by random values within a range of $[-40,\, 40]$ mm using a uniform distribution. 

Once the translation encoder is trained, we crop the resampled volume based on the translated mean shape. The cropped volume is then padded so that we cover enough area around the location to account for any changes in scale. Based on the dataset and a margin for safety, we crop the volume to a $512 \times 432 \times 352$ mm area. The cropped volume is then used to train the scale, rotation, and non-rigid encoders. Focusing on scale first and similar to the translation model, we perturb the trajectory at every episodic step. However, following \ac{MSL} principles, the  translation perturbations are smaller in magnitude than before to focus the search space on scale. Additionally, we also randomly pick between ``standard'' and ``fine-scale'' ranges of perturbations. When the latter is picked, it helps sample more poses closer to the convergence point when the model is at the final steps of its episode. This allows the model to better learn when to stop producing corrections, ensuring it will converge to a pose in inference. We also clip the scale parameters to lie between $[0.5,2]$ and impose a penalty for any scale predictions exceeding that range:
\begin{align}
    \max(0,s_{(.)}-2) + \max(0, 0.5-s_{(.)}) \textrm{,}
\end{align}
where $s_{(.)}$ is any of $s_{x}$, $s_{y}$, or $s_{z}$. This regularization threshold is determined based on the distribution of scales in the training set, which are computed during the rigid alignment coherent point drift step of Sec.~\ref{sec:sampling}.  The perturbation hyper parameters for training the scale encoder can be seen in Table~\ref{tbl:scale}. 
\begin{table}
\small

\begin{tabular}{|c|c|c|}
\hline
\textbf{Parameter} & \textbf{Standard Range} & \textbf{Fine Range} \\
\hline
Translation (mm)             & $[-16, 16]$ & $[-8, 8]$  \\
\hline
Scale         & $[0.7, 1.3]$ &  $[0.9, 1.1]$   \\
\hline

\end{tabular}
\caption{Perturbation ranges for training the scale encoder. The ``standard'' and ``fine'' ranges are randomly chosen at each step with $50\%$ probability each. Values are then randomly sampled within the resulting ranges using a uniform distribution.}
\label{tbl:scale}
\end{table}

We follow the same procedure to train rotation and non-rigid encoders. The perturbation hyper parameters for training rotation and non-rigid encoders can be seen in Tables \ref{tab:rotation} and \ref{tab:pca}, respectively. Finally, we also perform image-based data augmentations such as random affine transformations, random Gaussian noise, random brightness changes, and random intensity shifts for all pose encoders.
\begin{table}
\small

\begin{tabular}{|c|c|c|}
\hline
\textbf{Parameter} & \textbf{Standard Range} & \textbf{Fine Range} \\
\hline
Translation (mm)             & $[-12, 12]$ & $[-8, 8]$  \\
\hline
Scale         & $[0.97, 1.03]$ & $[0.99, 1.01]$   \\
\hline
Rotation (deg)         & $[-7.5, 7.5]$ & $[-4.5, 4.5]$   \\
\hline

\end{tabular}
\caption{Perturbation ranges for training the rotation encoder. The ``standard'' and ``fine'' ranges are randomly chosen at each step with $50\%$ probability each. Values are then randomly sampled within the resulting ranges using a uniform distribution.}
\label{tab:rotation}
\end{table}

\begin{table}
\small

\begin{tabular}{|c|c|c|}
\hline
\textbf{Parameter} & \textbf{Standard Range} & \textbf{Fine Range} \\
\hline
Translation (mm)             & $[-8, 8]$ & $[-4, 4]$  \\
\hline
Scale         & $[0.99, 1.01]$ & $[0.995, 1.015]$   \\
\hline
Rotation (deg)         & $[-4.5, 4.5]$ & $[-2.5, 2.5]$   \\
\hline

\end{tabular}
\caption{Perturbation ranges for training the non-rigid encoder. The ``standard'' and ``fine'' ranges are randomly chosen at each step with $50\%$ probability each. Values are then randomly sampled within the resulting ranges using a uniform distribution.}
\label{tab:pca}
\end{table}

\section{Training Local Refinement Model}
Once the non-rigid model is trained, we generate the non-rigid \ac{SDF} for each volume and in the original volume resolution. Because the \ac{SDF} output from the shape decoder is not necessarily a proper \ac{SDF}, we reinitialize it~\cite{sethian_1999} using the ITK software~\cite{mccormick2014itk}. To make the model robust to variations in the \ac{SDF}, prior to generating each \ac{SDF} we randomly add Gaussian noise with mean $0$ and standard deviation $0.01$ to the predicted latent vector.  We generate $10$ non-rigid \acp{SDF} for each volume this way, randomly choosing one \ac{SDF} to pair with the image when training.  

We follow the standard 3D U-Net architecture from \cite{cciccek20163d} and input both the original volume along with randomly picked non-rigid \ac{SDF} from the pool of \acp{SDF} for each volume. We further augment the \ac{SDF} channels through small rigid random affine transformations that are independent of the image channel. We weight the loss with $\lambda_{1}$ and $\lambda_{2}$ in (13) using the values mentioned in Table \ref{tab:refine}.
As mentioned in the main body, we only produce refinements within a narrow band of the non-rigid \acp{SDF} iso-boundary, which allows the model to focus only on the surface of the liver. We define the narrow band as being within $25$ mm of the iso-boundary. Finally, we set the $\rho$ in (13) to $12$ mm. We apply standard data augmentations to both input channels, such as random affine transformations (both channels), and random shifting intensity, random Gaussian noise, and random brightness shifts (only image channel) while training the local refinement model.

\begin{table}
    \centering
    \begin{tabular}{|c|c|}
    \hline
       Parameter & Value \\ \hline
        Batch Size & 2 \\
        Epochs & 150 \\
        Model Learning Rate & 0.001 \\
        Weight decay & 0.0001 \\
        $\lambda_{1}$ & 1 \\
        $\lambda_{2}$ & 0.1 \\

        \hline
    \end{tabular}
    \caption{Hyper-parameters for training local refinement model}
    \label{tab:refine}
\end{table}

\section{Competitor Methods}
We compare \ac{DIAS} with the competitor methods 2D PHNN~\cite{Raju_2020}, H-DenseUNet\cite{Li_2018}, nnU-Net (cascade)\cite{isensee2018nnu}, Adversarial learning\cite{yang_automatic_2017} for the liver segmentation task. We follow the same procedure including the data preprocessing, data augmentation and post processing as mentioned in their respective papers.
nnU-Net cascade preprocesses the image by resampling to the median voxel spacing. Third order spline interpolation is used for image data and nearest neighbour interpolation for label data. H-DenseUNet preprocesses the image by resampling to a fixed resolution $0.69 \times 0.69 \times 1 mm^{3}$ with the \ac{HU} values clipped to $[-200,250]$. For 2D PHNN, we use the original resolution and clip the \acp{HU} to $[-200, 250]$.
For Adversarial learning, we follow th same hyper parameter setting as H-DenseUNet.
For post processing, we select largest connected component for all the competitor methods coupled with 3D hole-filling. 
\section{Larynx CT Dataset}
The larynx dataset contains 142 non-contrast CT images from head and neck cancer patients. The larynx annotations are from the target delineation process during patients' radiotherapy treatment. We use the same one fold split as used in the ablation study of SOARS \cite{guo_organ_2020} to train and evaluate the performance of \ac{DIAS} and nnU-Net, which includes 89, 17 and 36 training, validation and testing patients. Similar to \ac{MSD} dataset, we resample the volume to $4 \times 4 \times 1.5$ mm resolution and pad across the dimensions to create an unified volume of size of physical dimension $512 \times 512 \times 612$ mm. We further crop the volume after translation model to a $432 \times 284 \times 156$ mm to train scale, rotation and non-rigid encoders.

\section{Ablation study}
\ac{DIAS} simply cannot converge without \ac{MSL} or an agent-based formulation, so these components cannot be ablated. However, we did measure the performance after each stage of \ac{MSL}. In addition, we experimented with different numbers of PCA loadings and we can find that the model achieves good performance with even 90$\%$ explained variance, but that performance is slightly better with 72\% explained variance (see Table \ref{tbl:MSD_ABLATION_LIVER}).
\begin{table}[t]
\centering
\footnotesize

\begin{tabular}{|l|l|l|l|l|l|l|l|}
\hline
\multicolumn{2}{|c|}{Model} & \multicolumn{2}{c|}{DSC (\%)} & \multicolumn{2}{c|}{ASSD (mm)} & \multicolumn{2}{c|}{HD (mm)} \\  \hline
\multicolumn{2}{|l|}{Trans.}               & \multicolumn{2}{l|}{79.2 $\pm$ 12.3 }      &\multicolumn{2}{l|}{7.1 $\pm$ 2.4 }  &\multicolumn{2}{l|}{54.1 $\pm$ 23.6 } \\
\multicolumn{2}{|l|}{Trans. + scale}               & \multicolumn{2}{l|}{83.4 $\pm$ 12.1 }      &\multicolumn{2}{l|}{5.8 $\pm$ 2.2}  &\multicolumn{2}{l|}{47.0 $\pm$ 17.4} \\
\multicolumn{2}{|l|}{Trans. + scale + rot.}               & \multicolumn{2}{l|}{89.1 $\pm$ 10.5 }      &\multicolumn{2}{l|}{4.3 $\pm$ 2.6}  &\multicolumn{2}{l|}{31.6 $\pm$ 18.1} \\
\multicolumn{2}{|l|}{w/o refine, pca=90\%}               & \multicolumn{2}{l|}{95.9 $\pm$ 1.2 }      &\multicolumn{2}{l|}{1.8 $\pm$ 1..1}  &\multicolumn{2}{l|}{23.7 $\pm$ 13.8} \\
\multicolumn{2}{|l|}{w/o refine, pca=72\%}               & \multicolumn{2}{l|}{96.1 $\pm$ 0.9 }      &\multicolumn{2}{l|}{1.5 $\pm$ 0.7}  &\multicolumn{2}{l|}{23.4 $\pm$ 12.2} \\
\multicolumn{2}{|l|}{DISSM w refine}               & \multicolumn{2}{l|}{96.5 $\pm$ 0.7 }      &\multicolumn{2}{l|}{1.1 $\pm$ 0.7}  &\multicolumn{2}{l|}{21.4 $\pm$ 11.8} \\
\hline
\end{tabular}

\caption{Ablation study on MSD liver dataset. "Trans." refers to translation, "rot." refers to rotation.}
\label{tbl:MSD_ABLATION_LIVER}
\end{table}

\begin{figure*}
   \centering
     \setlength\tabcolsep{0pt}
     \renewcommand{\arraystretch}{0}
    \begin{tabular}{cccccc}
    2D PHNN  &  nnU-Net & H-DenseUNet & Adv. learning & \acs{DIAS} w/o refine & \acs{DIAS} w refine \\
    \includegraphics[width=3cm]{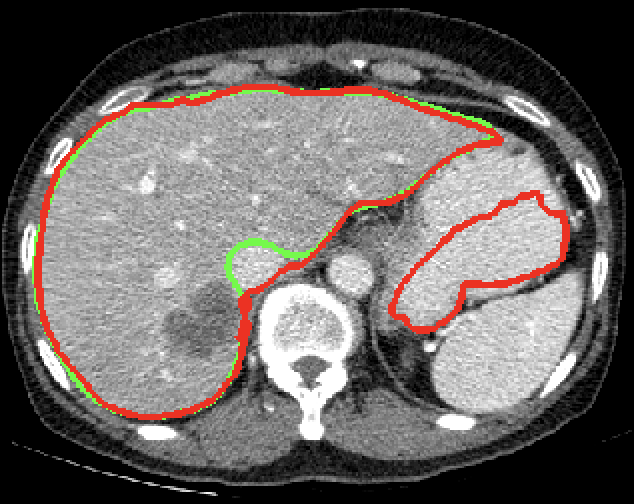} & \includegraphics[width=3cm]{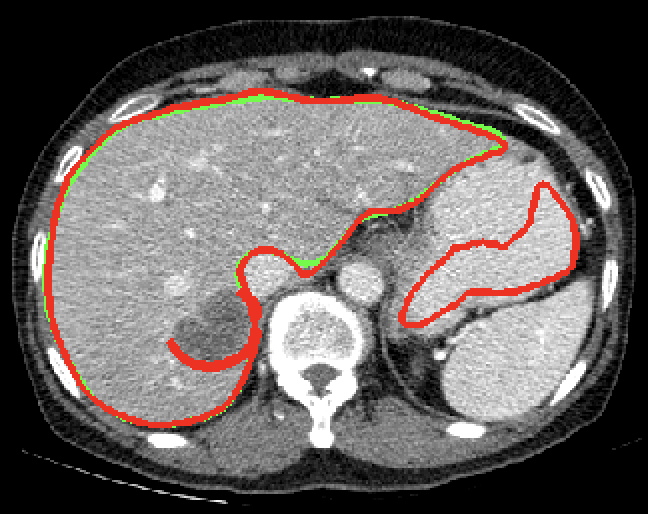} & \includegraphics[width=3cm]{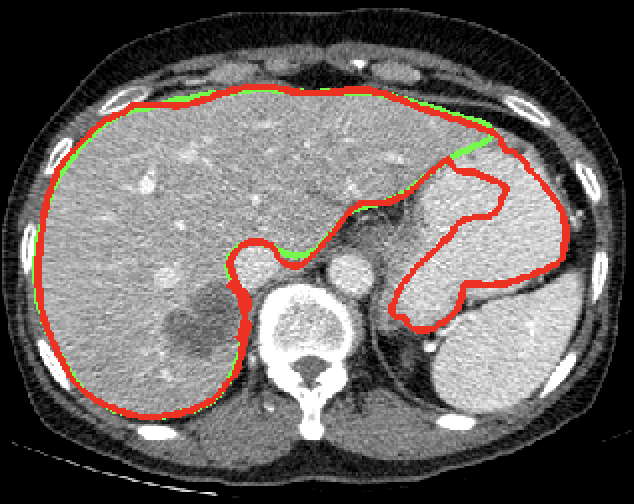} &
     \includegraphics[width=3cm,height=2.39cm]{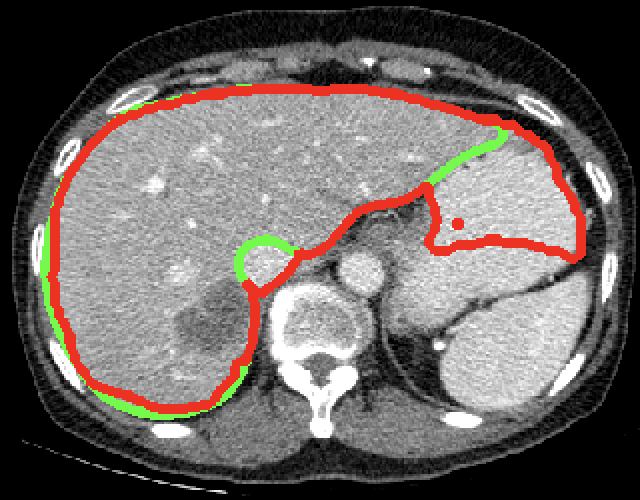} &
       \includegraphics[width=3cm]{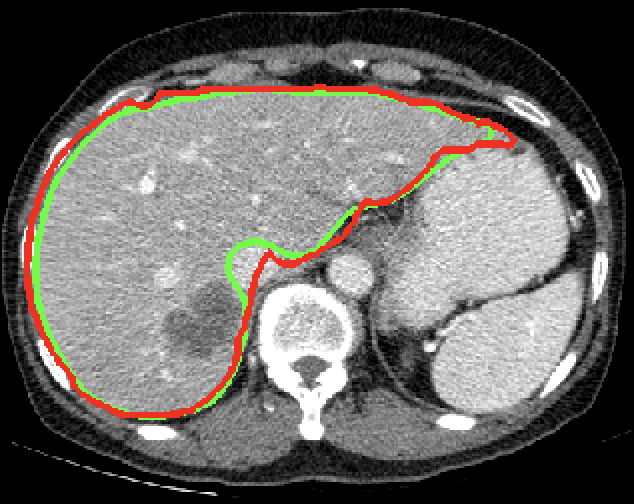} &
     \includegraphics[width=3cm]{images/liver_70_refine.png} \\
     
        \includegraphics[width=3cm]{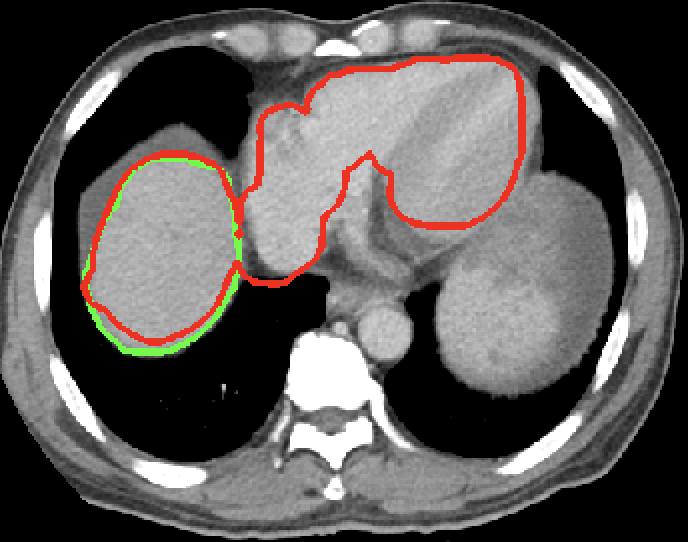} & \includegraphics[width=3cm]{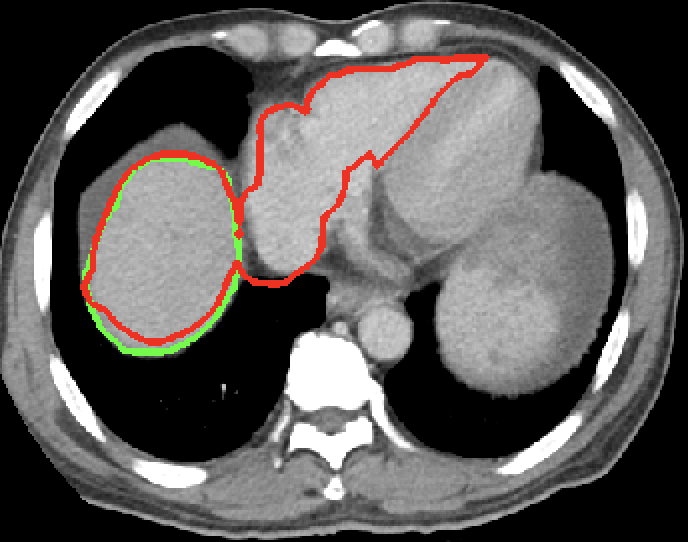} & \includegraphics[width=3cm]{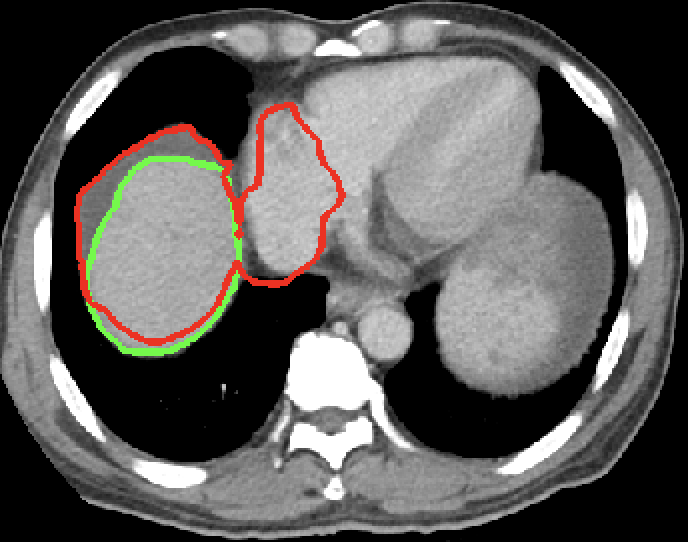} &
     \includegraphics[width=3cm,height=2.364cm]{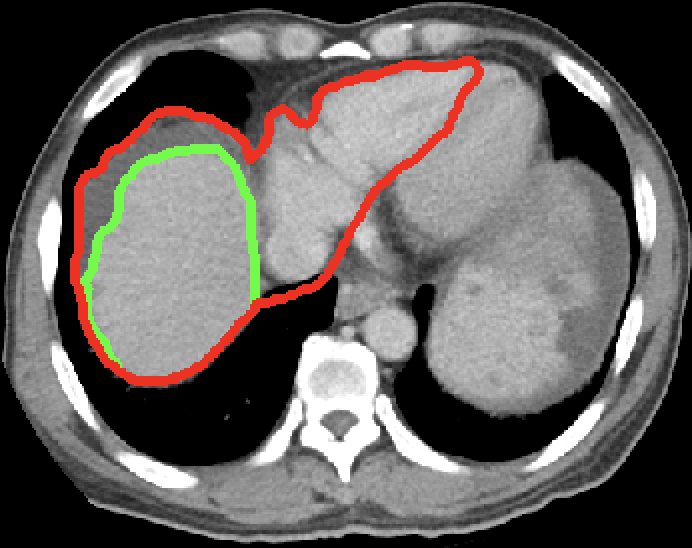} & 
     \includegraphics[width=3cm]{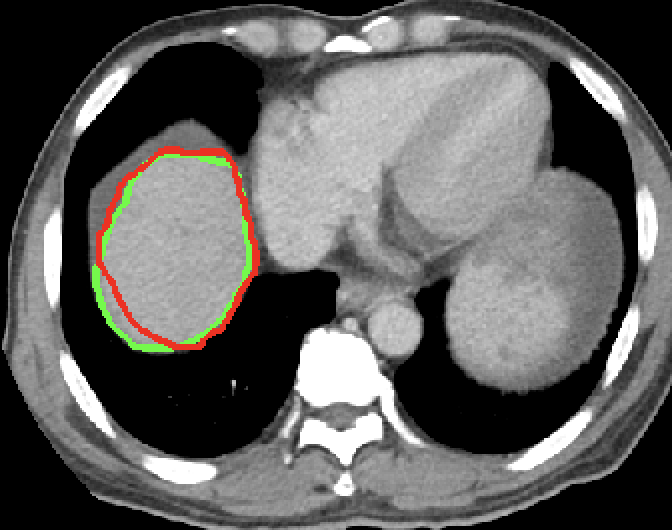} & 
     \includegraphics[width=3cm]{images/cghm_02097_refine1.png} \\
     
     \includegraphics[width=3cm]{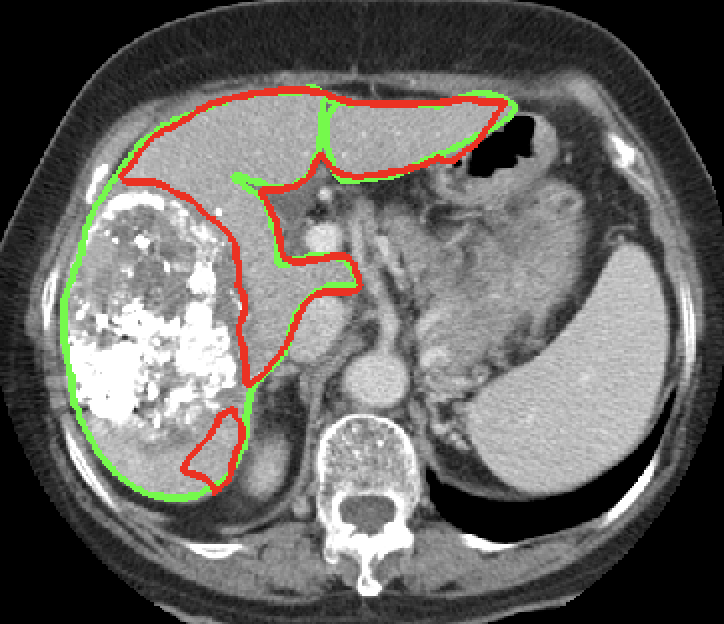} & \includegraphics[width=3cm]{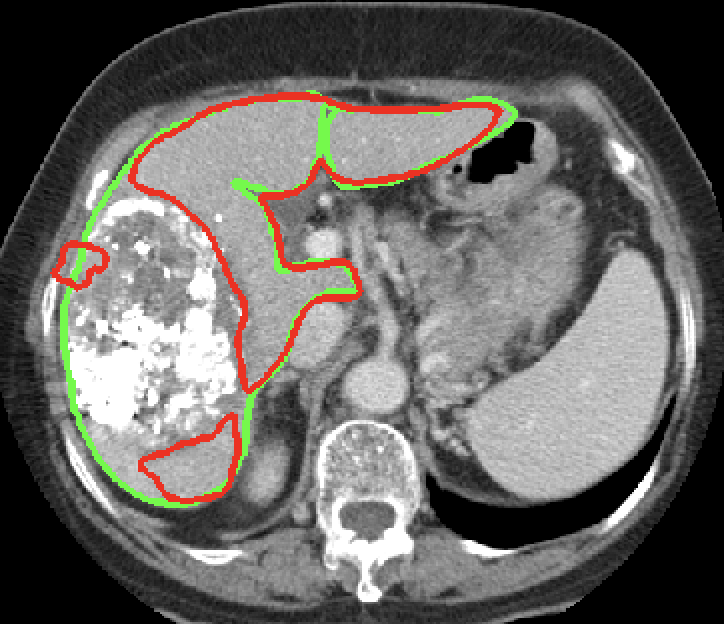} & \includegraphics[width=3cm]{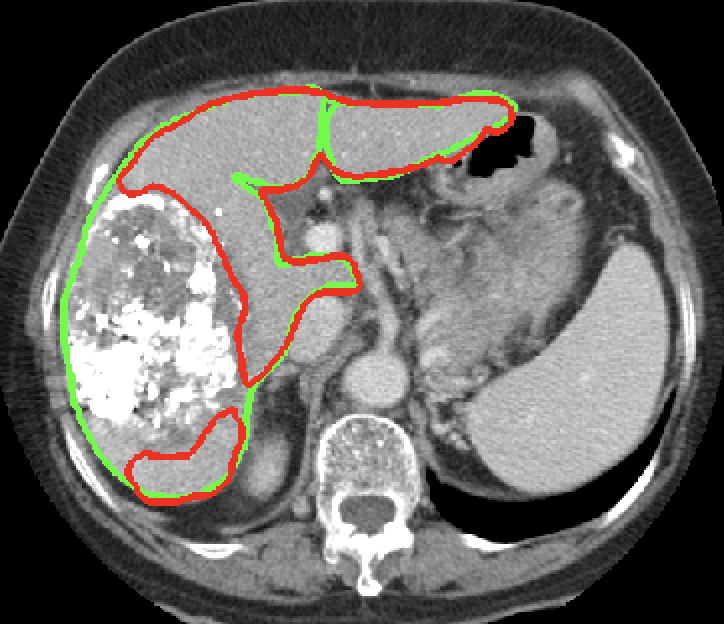} &
     \includegraphics[width=3cm,height=2.59cm]{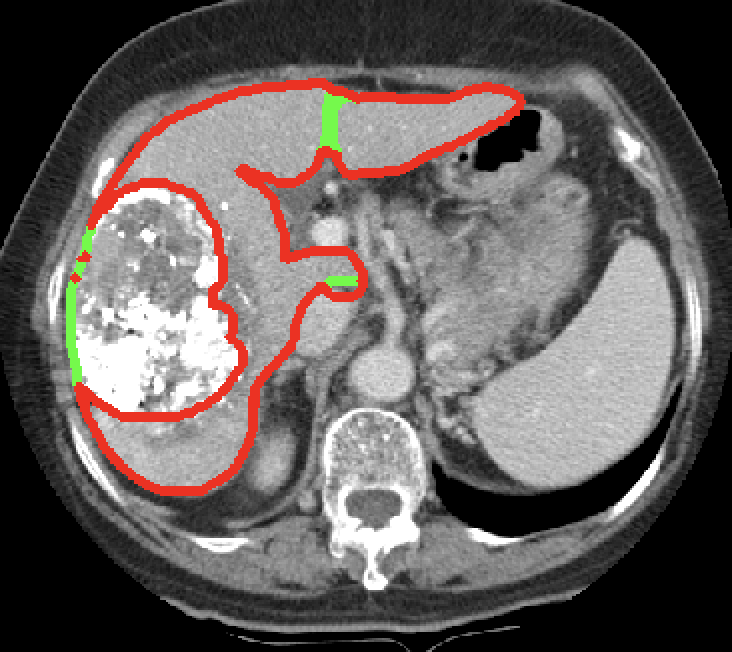} &
     \includegraphics[width=3cm]{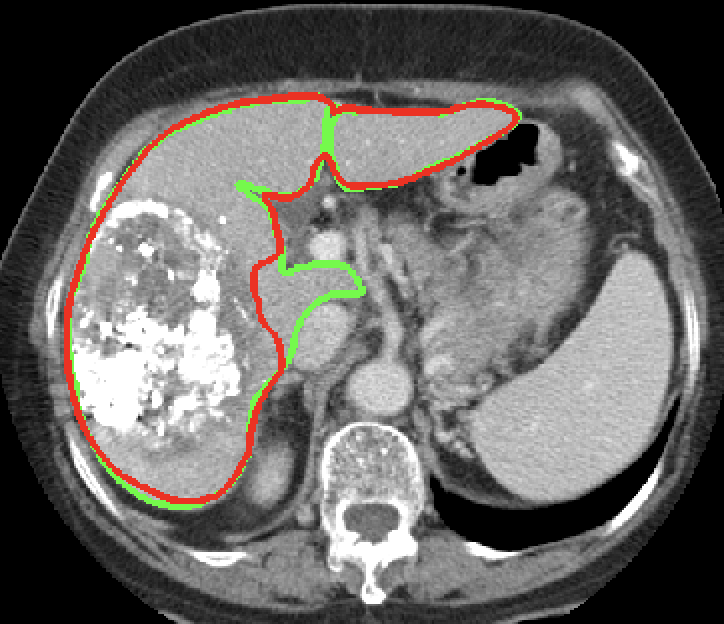} &
     \includegraphics[width=3cm]{images/cgmh_02248_refine.png} \\
     
     \includegraphics[width=3cm]{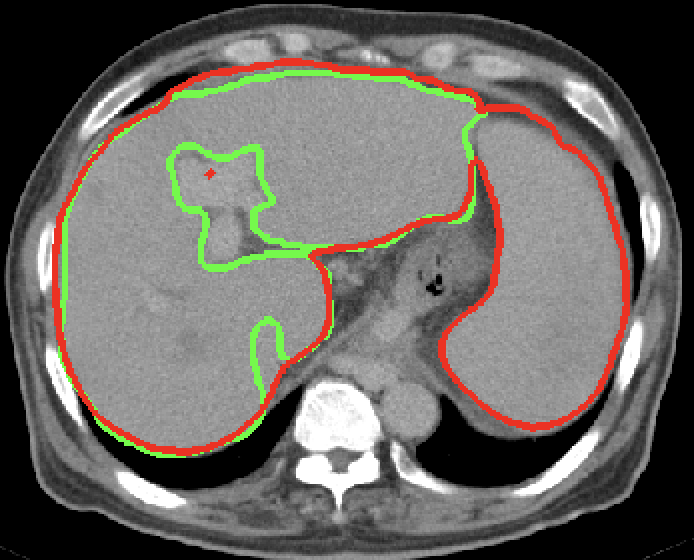} & \includegraphics[width=3cm]{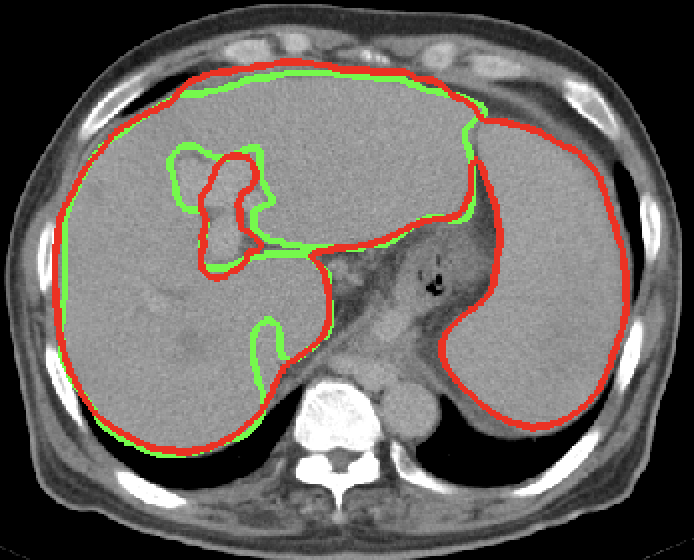} & \includegraphics[width=3cm]{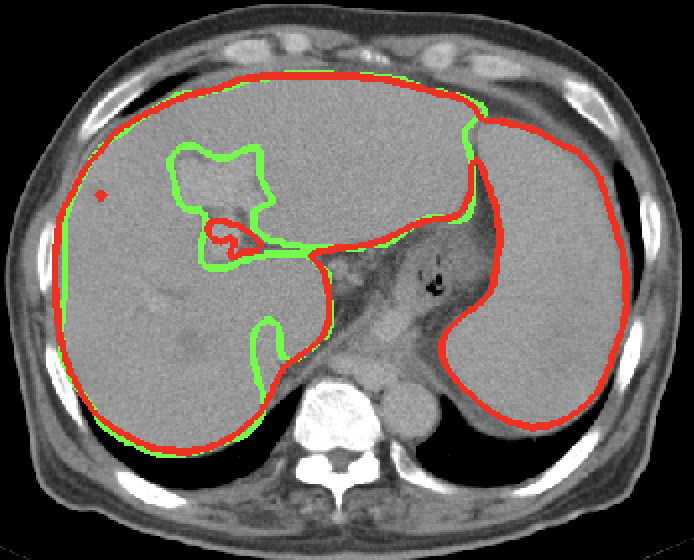} &
     \includegraphics[width=3cm,height=2.42cm]{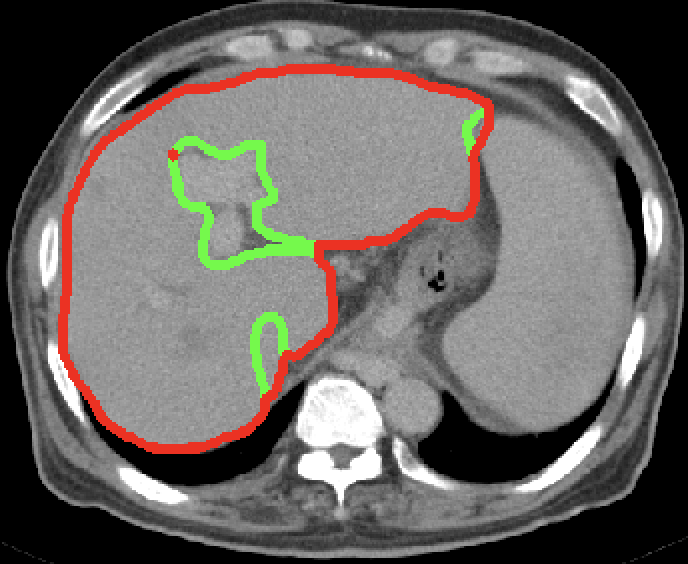} &
     \includegraphics[width=3cm]{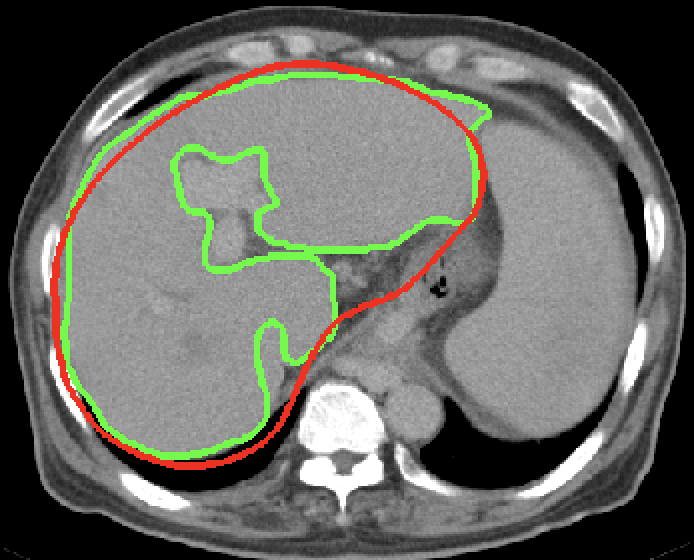} &
     \includegraphics[width=3cm]{images/cgmh_06846_refine.png}  
    \end{tabular}
    \renewcommand{\arraystretch}{1}
    \caption{Qualitative comparison of \acs{DIAS} (without or with refinement) versus four state-of-the-art competitor approaches \cite{Harrison_2017,isensee_nnu-net_2021,Li_2018,yang_automatic_2017}. Red and green contours represent the current estimate and the ground truth of liver delineations, respectively. The first and second-to-fourth rows are drawn from the \ac{MSD} and clinical dataset, respectively. Fourth row is worst-case performance for \ac{DIAS}.} 
    \label{fig:qualitative}
\end{figure*}

\begin{figure*}
   \centering
     \setlength\tabcolsep{0pt}
     \renewcommand{\arraystretch}{0}
    \begin{tabular}{cccc}
        SOARS & nnU-Net &  \acs{DIAS} w/o refine & \acs{DIAS} w refine \\
     \includegraphics[width=3cm]{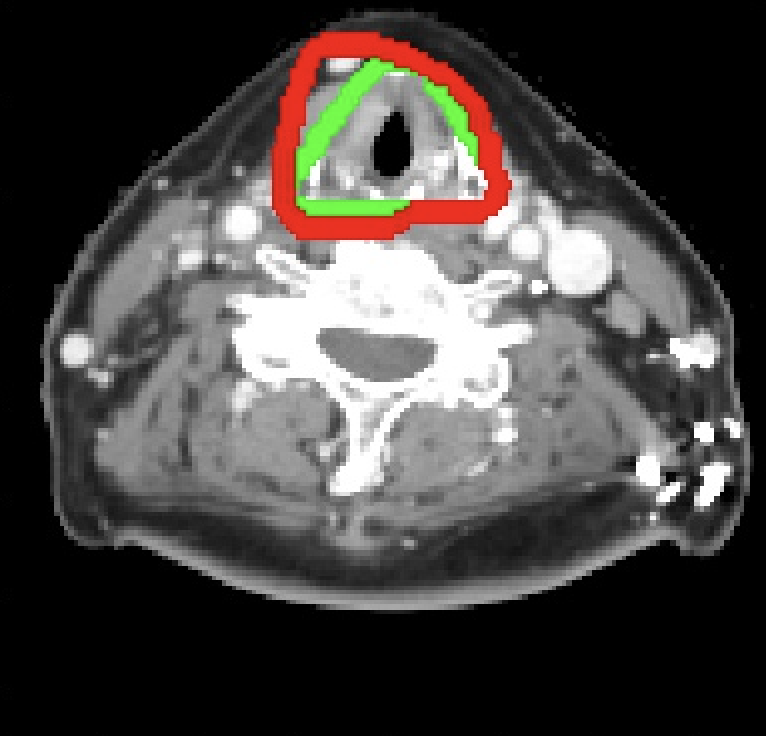} & \includegraphics[width=3cm]{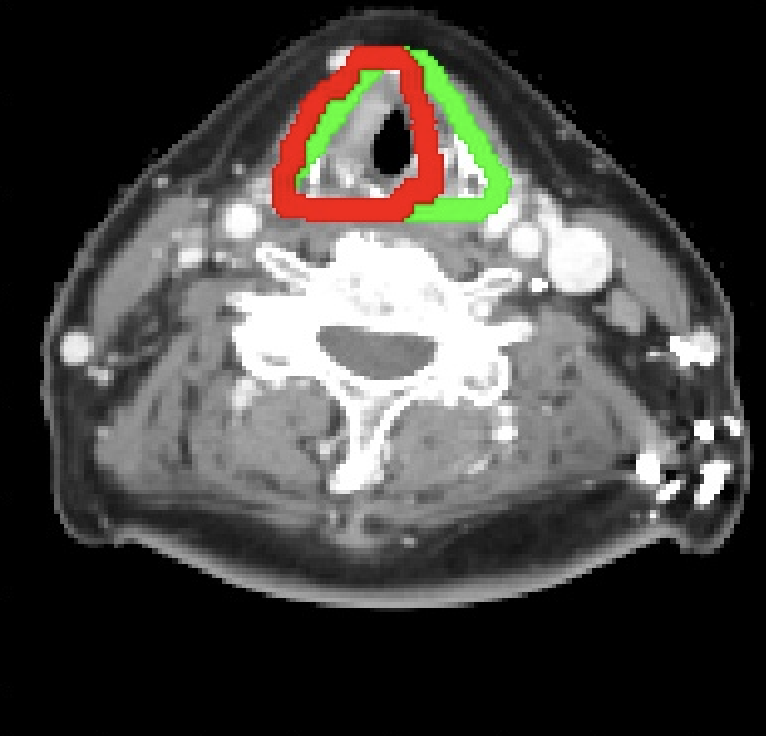} &
     \includegraphics[width=3cm]{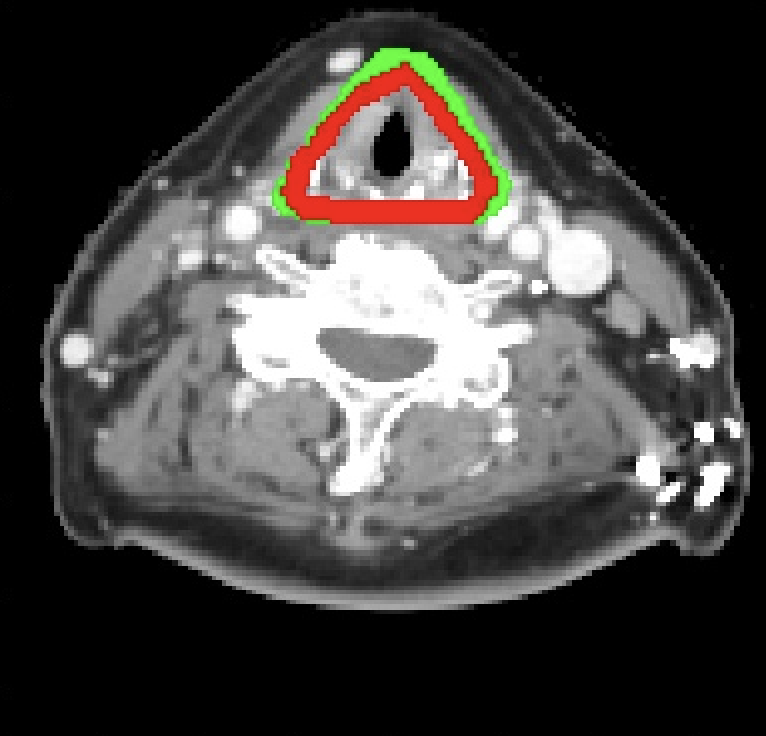} &
     \includegraphics[width=3cm]{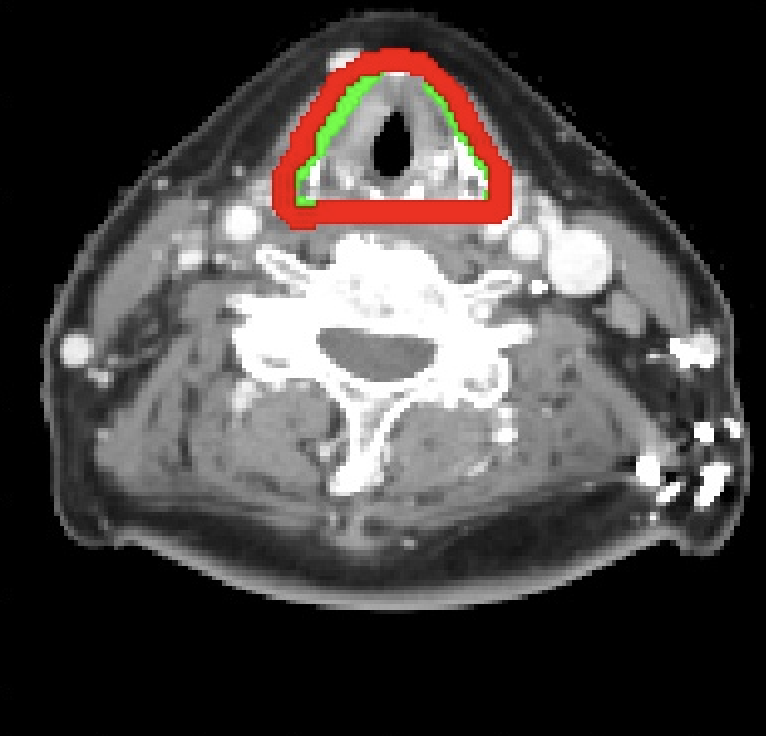} \\
     
      \includegraphics[width=3cm,height=2.88cm]{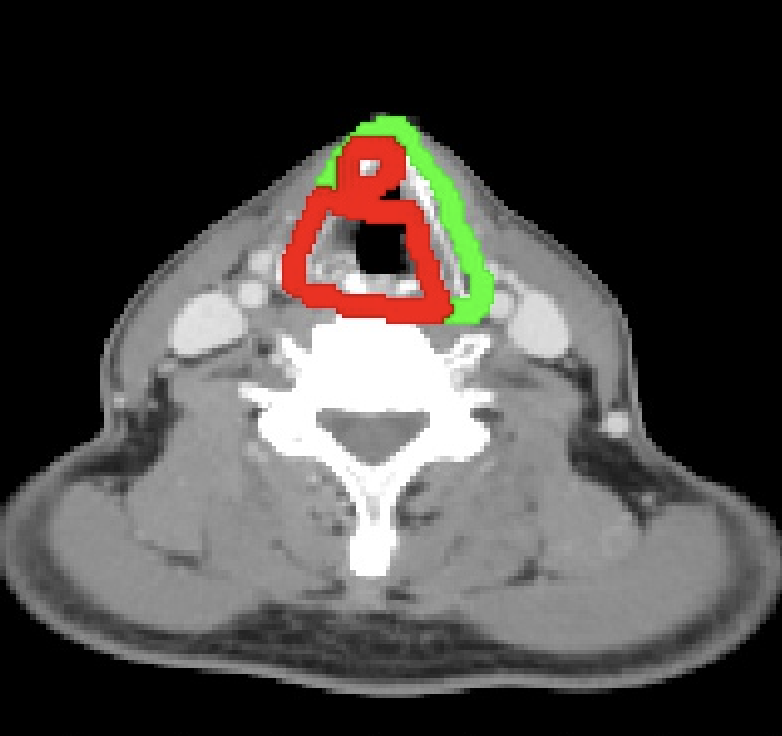} & \includegraphics[width=3cm, height=2.88cm]{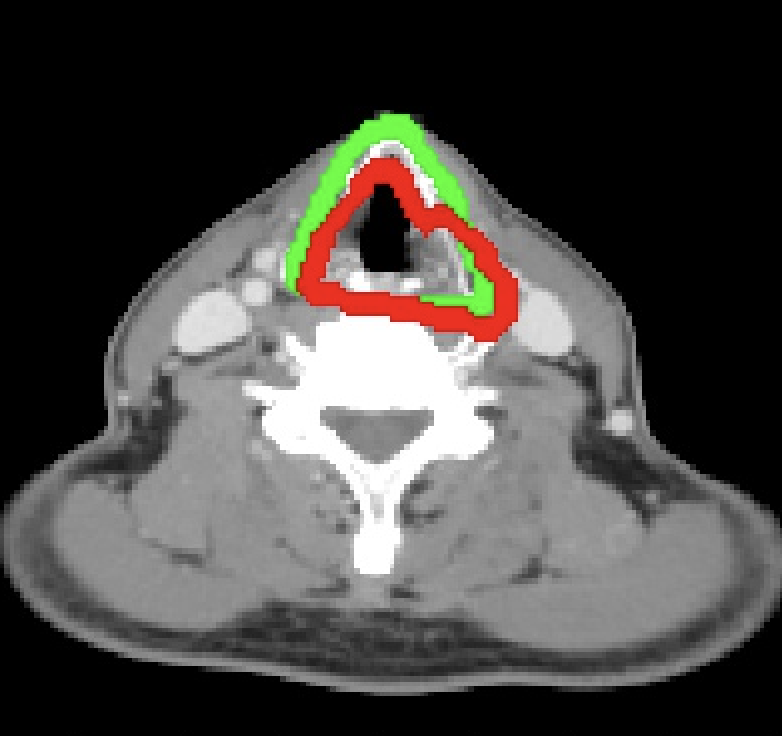} &
     \includegraphics[width=3cm, height=2.88cm]{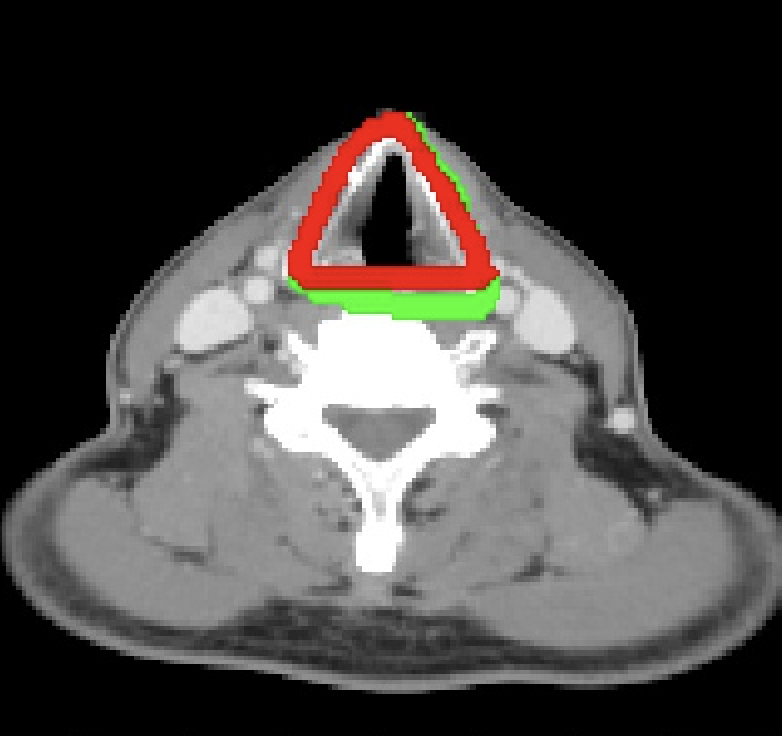} &
     \includegraphics[width=3cm, height=2.88cm]{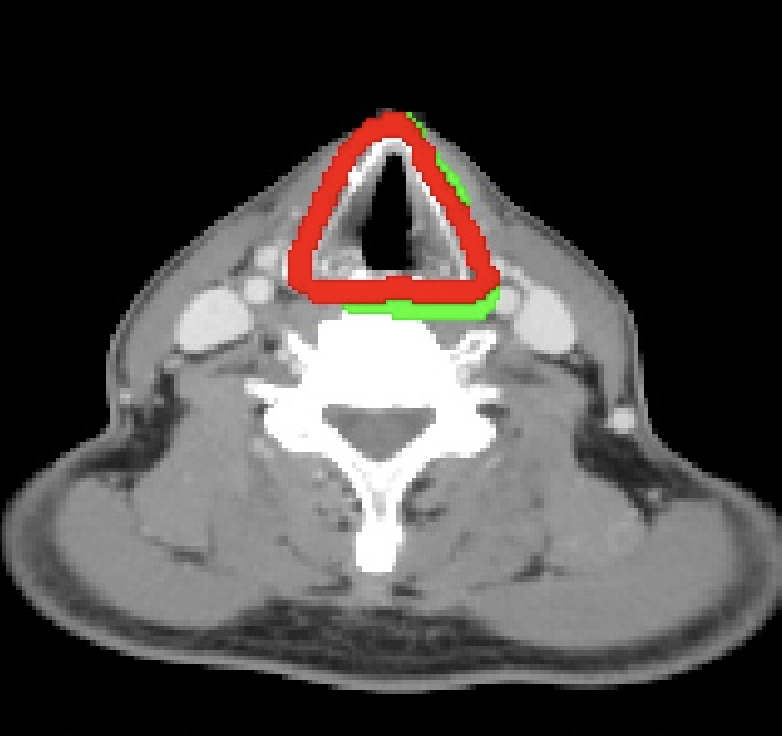} \\
     
      \includegraphics[width=3cm, height=2.88cm]{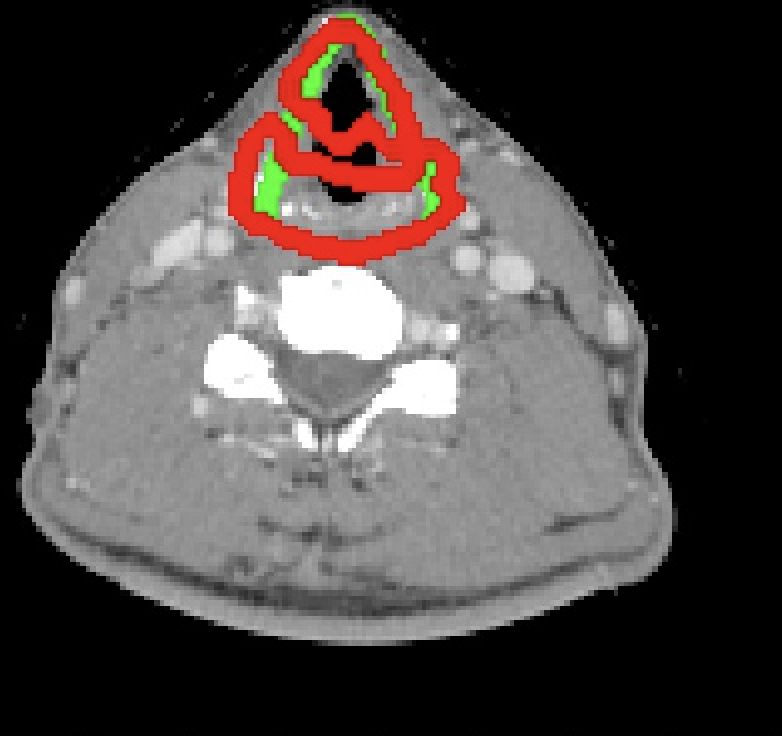} & \includegraphics[width=3cm,height=2.88cm]{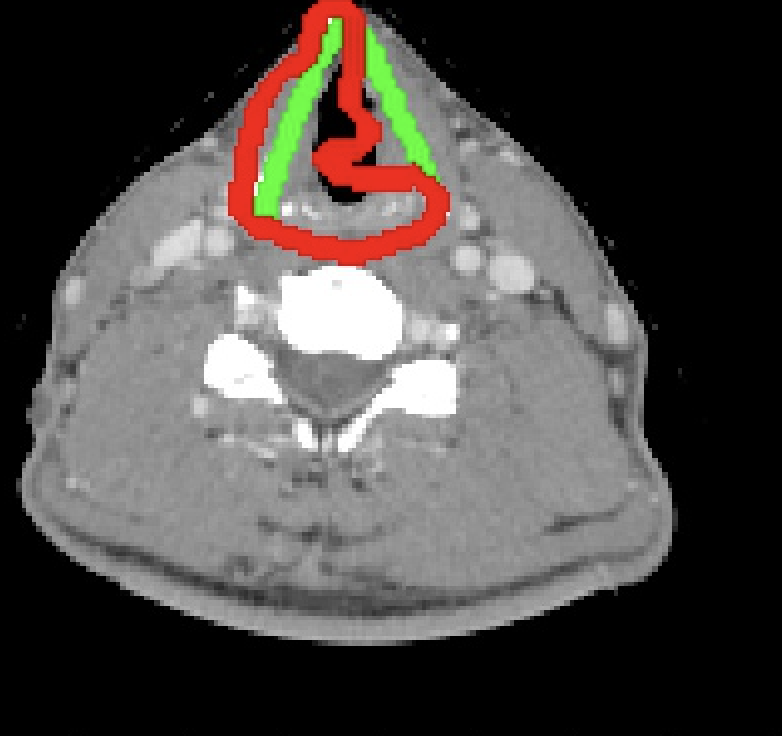} &
     \includegraphics[width=3cm,height=2.88cm]{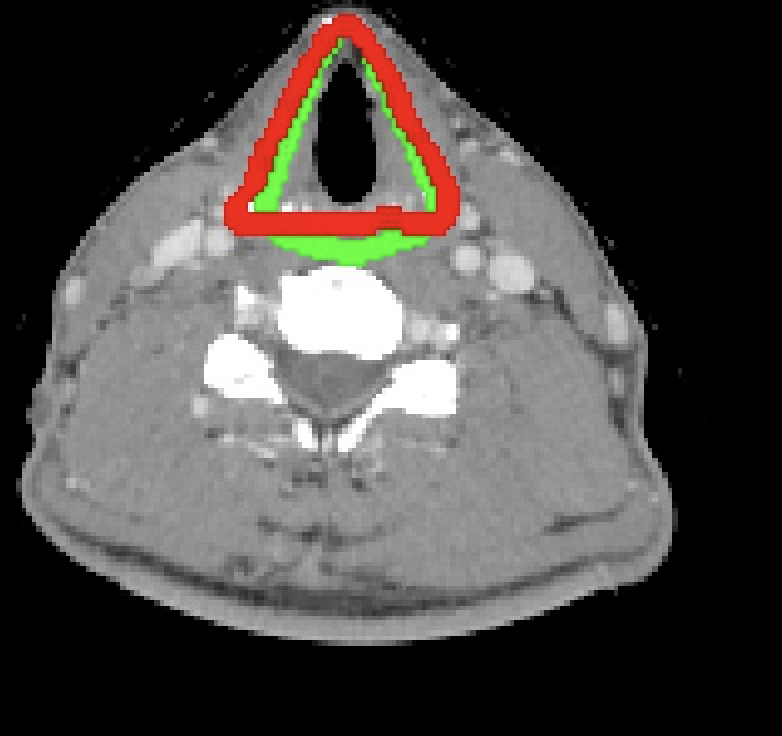} &
     \includegraphics[width=3cm, height=2.88cm]{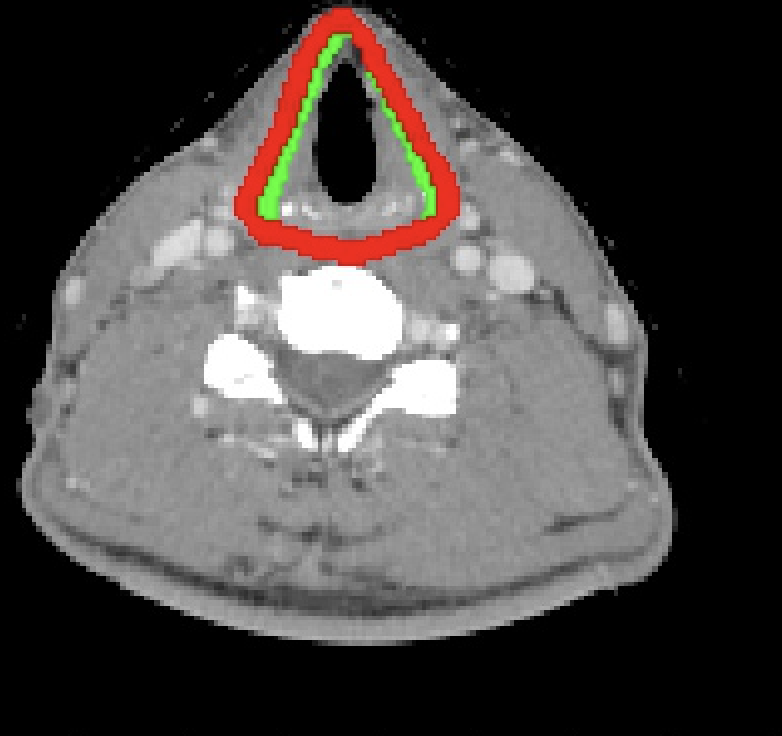} \\
      
    \end{tabular}
    \renewcommand{\arraystretch}{1}
    \caption{Qualitative comparison of \acs{DIAS} on larynx dataset. Red and green contours represent the current estimate and the ground truth of larynx delineations, respectively. } 
    \label{fig:qualitative_larynx}
\end{figure*}

\newpage 
\section{Qualitative examples}
\textbf{Liver:} First two rows in Figure \ref{fig:qualitative} shows how \ac{DIAS} does not recognise the heart whereas other competitor methods recognizes heart as a part of liver. In second row, \ac{DIAS} recognizes \ac{TACE} has a part of liver region whereas other competitor methods recognizes \ac{TACE} as part of background. In the last row, \ac{DIAS} does note recognize a large spleen as liver whereas the competitor methods do recognize spleen as a part of liver.
\textbf{Larynx:} Competitor methods does not recognize the shape of the larynx as it is merely a pixel-level segmentation whereas \ac{DIAS} recognizes the shape of larynx.

\end{document}